%% file: main.tex
\documentclass{article} 
\usepackage{iclr2026_conference,times}

\input{math_commands.tex}

\usepackage{hyperref}
\usepackage{url}

\usepackage{graphicx}
\usepackage{booktabs}
\usepackage[dvipsnames]{xcolor}  
\usepackage[most]{tcolorbox}
\usepackage{chemformula} 
\usepackage{multirow}
\usepackage{makecell}

\hypersetup{
        final,
        colorlinks,
        linkcolor={BrickRed},
        citecolor={blue},
        urlcolor={DarkOrchid}
        }
\usepackage{titletoc}    
\usepackage{etoolbox}    
\usepackage{tocloft}     

\RequirePackage{xspace}
\makeatletter
\DeclareRobustCommand\onedot{\futurelet\@let@token\@onedot}
\def\@onedot{\ifx\@let@token.\else.\null\fi\xspace}

\def\ie{\emph{i.e}\onedot}

\def\etal{\emph{et al}\onedot}

\usepackage{tikz}
\DeclareRobustCommand{\atomdot}[1]{%
  \tikz[baseline=-0.6ex]\node[circle, fill=#1, draw=black, inner sep=2pt] {};%
}
\DeclareRobustCommand{\hydrogendot}{%
\tikz[baseline=-0.6ex]%
\node[circle, fill=white, draw=black, inner sep=2pt] {};
}
\title{Latent Diffusion-Based 3D Molecular Recovery from Vibrational Spectra}

\author{Wenjin Wu\textsuperscript{$\dagger$}, Ale\v{s} Leonardis\textsuperscript{$\dagger$}, Linjiang Chen\textsuperscript{$\dagger\ddagger\S$}, Jianbo Jiao\textsuperscript{$\dagger$}
\\
\textsuperscript{$\dagger$} School of Computer Science, University of Birmingham\\
\textsuperscript{$\ddagger$} School of Chemistry, University of Birmingham\\
\textsuperscript{$\S$} State Key Laboratory of Precision and Intelligent Chemistry, \\
University of Science and Technology of China\\
\texttt{wxw354@student.bham.ac.uk}, \\
\texttt{\{a.leonardis, j.jiao\}@bham.ac.uk}, \\
\texttt{linjiangchen@ustc.edu.cn}
}

\iclrfinalcopy 
\begin{document}

\maketitle
\begin{abstract}
Infrared (IR) spectroscopy, a type of vibrational spectroscopy, is widely used for molecular structure determination and provides critical structural information for chemists. 
However, existing approaches for recovering molecular structures from IR spectra typically rely on one-dimensional SMILES strings or two-dimensional molecular graphs, which fail to capture the intricate relationship between spectral features and three-dimensional molecular geometry.
Recent advances in diffusion models have greatly enhanced the ability to generate molecular structures in 3D space. 
Yet, no existing model has explored the distribution of 3D molecular geometries corresponding to a single IR spectrum.
In this work, we introduce \texttt{IR-GeoDiff}, a latent diffusion model that recovers 3D molecular geometries from IR spectra by integrating spectral information into both node and edge representations of molecular structures.
We evaluate IR-GeoDiff from both spectral and structural perspectives, demonstrating its ability to recover the molecular distribution corresponding to a given IR spectrum. 
Furthermore, an attention-based analysis reveals that the model is able to focus on characteristic functional group regions in IR spectra, qualitatively consistent with common chemical interpretation practices.

\centering Project page: \url{https://wenjin886.github.io/IR-GeoDiff/}
\end{abstract}

\section{Introduction}

Chemists have developed various molecular spectroscopic methods to elucidate molecular structures. 
Different spectroscopy contains different molecular information, and each individual spectroscopy can be seen as a dimensionally reduced representation of the full 3D molecular structures.
Among these molecular spectra, infrared (IR) spectroscopy is widely applied in different scenarios~\citep{irfood, tinetti2013spectroscopy, muro2015vibrational, renner2019data} and is capable of identifying various types of compounds~\citep{becucci2016high, hadjiivanov2020power, adhikary2017transparent}. 
Directly recovering molecular structures from infrared spectra holds significant practical value. If highly automated structural interpretation becomes feasible, it could greatly accelerate molecular screening and validation workflows in real-world applications such as materials design~\citep{lansford2020infrared}. 
IR spectra originate from molecular vibrations, and peaks appearing in specific spectral regions correspond to particular vibrational modes. Chemists typically interpret IR spectra by identifying characteristic bonds or functional groups based on empirical rules~\citep{coates2000interpretation} or through visual comparisons with reference spectra from standard databases~\citep{zhou2012spectral} or computational simulations~\citep{henschel2020theoretical,esch2021quantitative}. 
However, IR spectra often exhibit complex peak patterns, especially in the fingerprint region (400-1,350 cm\textsuperscript{-1}), posing significant interpretation challenges. Furthermore, limited database coverage and variability of experimental conditions constrain the effectiveness of visual comparison methods.

Researchers have applied artificial neural networks to elucidate IR spectra of small organic molecules since the 1990s~\citep{gasteiger1993neural,burns1993feed, hemmer1999deriving}, the primary task of which was classifying predefined functional groups~\citep{gasteiger1993neural,burns1993feed,judge2008sensitivity, fine2020spectral, wang2020functional, enders2021functional, jung2023automatic,wang2023infrared, lu2024patch, rieger2023understanding}, the specific groups of atoms bonded in certain arrangements that together exhibit their own characteristic properties.
Traditionally, researchers have applied various machine learning models, including multi-layer perceptrons (MLPs)~\citep{fine2020spectral}, support-vector machines (SVMs)~\citep{wang2020functional}, and convolutional neural networks (CNNs)~\citep{enders2021functional,jung2023automatic,wang2023infrared,lu2024patch, rieger2023understanding}, to classify specific substructures (\ie, functional groups) within molecules based on IR spectra.

Regarding the direct prediction of entire molecular structures from IR spectra, Hemmer \etal~\cite{hemmer1999deriving}~proposed a method to derive 3D molecular structures from IR spectra by representing molecules as radial distribution function (RDF) codes. However, their method depended on database retrieval to reconstruct molecular geometries from RDF codes. After this early attempt, the task of directly predicting molecular structures received limited attention until recently. 
Ellis \etal~\cite{ellis2023deep}~revisited this problem, developing a deep Q-learning-based model for 2D molecular graph generation. More recently, Transformer-based methods have introduced substantial advances, directly generating molecular structures from IR spectra along with chemical formulas, and demonstrating remarkable performance on both simulated and experimental datasets~\citep{alberts2024leveraging,wu2025transformer,alberts2025setting}.
In these studies, molecular structures are represented as 1D representations in terms of SMILES (Simplified Molecular Input Line Entry System)~\citep{weininger1988smiles} strings. 
Although SMILES strings well encode structural information through developed grammar rules, they do not provide explicit interatomic relationships. 
In addition, a single molecular structure can correspond to multiple valid SMILES strings, which introduces representational ambiguity and requires models to implicitly learn the underlying grammar.
Moreover, no matter whether molecules are represented as 1D SMILES strings or 2D molecular graphs, essential 3D structural information is inevitably lost or severely reduced, despite IR spectra originating from molecular vibrations, which are inherently 3D phenomena reflecting atomic spatial arrangements. This representation gap makes it difficult to understand how the model interprets spectral features accurately. However, using 3D geometries to represent molecular structure requires the model to have a deeper understanding of molecular structure to complement 3D spatial information and generate valid molecules.

Recently, diffusion models (DMs)~\citep{sohl2015deep,ho2020denoising} have attracted significant attention due to their remarkable performance in generative tasks across diverse fields, including computer vision~\citep{ho2020denoising,rombach2022high}, natural language processing~\citep{austin2021structured,hoogeboom2021argmax}, and molecular generation in both 2D~\citep{vignac2023digress,jo2022score} and 3D~\citep{hoogeboom2022equivariant,xu2022geodiff,wu2022diffusion,xu2023geometric,huang2023mdm,o20233d,qiang2023coarse} space. Hoogeboom \etal~\cite{hoogeboom2022equivariant}~first introduced E(3)-equivariant diffusion models (EDMs) for molecular generation tasks in 3D space. Building upon this work, Xu \etal~\cite{xu2023geometric}~proposed geometric latent diffusion models (GEOLDM), utilising latent space representations to model complex molecular geometries.
In the molecular domain, desired conditions typically are chemical properties, such as polarizability and orbital energies, thus simply concatenating node features can realise conditional generation~\citep{hoogeboom2022equivariant, xu2023geometric}.
More complex conditions for drug design, such as protein pockets~\citep{schneuing2024structure, guan2023d, peng2022pocket2mol} and reference molecular structures~\citep{chen2023shapeconditioned, adams2025shepherd}, often require either customised neural network architectures or specialised sampling strategies~\citep{peng2022pocket2mol, ayadi2024unified} to effectively incorporate conditional information into diffusion models.
Recently, Cheng \etal~\cite{cheng2024determining}~proposed KREED, which determines 3D molecular structures from rotational spectra by leveraging the molecular formula, Kraitchman’s substitution coordinates, and principal moments of inertia. Bohde \etal~\cite{bohde2025diffms}~introduced DiffMS, which reconstructs molecules represented as 2D graphs from mass spectra, given the molecular formula.
Despite this progress, there is currently no diffusion model that is conditioned on full IR spectra and learns a distribution over three-dimensional molecular geometries. Likewise, no evaluation protocol has been established for this new spectrum-to-geometry recovery task, which is fundamentally different from de novo molecular generation or conformer generation. The objective is not to encourage diversity in the generated molecules, but to recover geometries that are consistent with a given spectrum and to reduce the candidate space as much as possible. 

To address these challenges, we propose a new \textbf{IR} spectra-guided \textbf{Geo}metric latent \textbf{Diff}usion model (\texttt{IR-GeoDiff}) to recover the distribution of molecular geometries represented by IR spectroscopy. 
As molecular vibrations primarily affect the interatomic distance, we not only incorporate IR spectral information into atomic features, but also the edges between any two atoms within the molecule. 
Furthermore, by visualising the edge-spectrum and atom-spectrum cross-attention maps,
we found that the way in which our model analyses spectra and structure aligns with the quantum theories of IR spectroscopy. 
In summary, our contributions are as follows:

\begin{enumerate}
    \item We introduce a new task of \textit{recovering the distribution of 3D molecular geometries from infrared spectroscopy}. This formulation bridges molecular structure generation and spectroscopic analysis, and highlights the intrinsic link between vibrational spectral patterns and 3D molecular geometries.
    \item To the best of our knowledge, the proposed \texttt{IR-GeoDiff} is the first model to directly recover 3D molecular geometries from 1D infrared spectra, introducing a new paradigm for the automatic interpretation of IR spectra by leveraging 3D diffusion models.
    \item We propose a comprehensive set of evaluation metrics that assess whether the recovered molecular structures are consistent with the distribution encoded in the input infrared spectra, from both structural and spectral perspectives.
    \item We further investigate how the proposed model works, 
    and find that the model is able to focus on spectral regions associated with characteristic functional groups. This behaviour qualitatively resembles how chemists interpret IR spectra, providing preliminary insights into the model’s interpretability.
\end{enumerate}

\section{Methods}

\subsection{Preliminaries}
\paragraph{Problem definition.}
\label{sec: prob-define}
For a molecule with $N$ atoms, the molecular geometry is represented as $G = \langle \mathbf{x}, \mathbf{h} \rangle$, where $\mathbf{x} = (\mathbf{x}_1, \dots, \mathbf{x}_N) \in \mathbb{R}^{N \times 3}$ denotes the 3D Cartesian coordinates of atoms, and $\mathbf{h} = (h_1, \dots, h_N) \in \mathbb{R}^{N \times 1}$ denotes their atomic types. 
Denote the input infrared spectrum as $S \in \mathbb{R}^{1 \times l}$, where $l$ is the number of sampled spectral points.
Our goal is to learn a probabilistic model $\theta$ that captures the conditional distribution of molecular geometries given an IR spectrum, \ie $p_\theta(G|S)$. Unlike conventional conditional molecular generation tasks, which aim to produce diverse structures under certain constraints like polarizability and protein pockets, we focus on recovering the set of 3D molecular geometries consistent with a given IR spectrum, where the goal is not to encourage diversity, but to reduce the candidate space as much as possible. 

In principle, the IR spectrum may correspond to a unique molecular structure, and the conditional distribution could collapse to a single geometry.
Although an observed IR spectrum in common real experimental conditions is a superposition of contributions from many different conformers, combined through Boltzmann weighting~\citep{mcgill2021predicting}, the simulated experimental spectrum in computational chemistry is typically obtained as a Boltzmann average of over conformer-specific spectra~\citep{marton2023artificial, yurenko2007many}. Every single, distinct conformer in that ensemble theoretically corresponds to its own unique IR spectrum~\citep{von2008mid}.
Thus, the problem that this work aims to address is this foundational ``one-to-one" theoretical mapping: recovering a single geometry from its corresponding theoretical spectrum. We argue that solving this task is a necessary and critical first step before tackling the more complex problem of deconvolving experimental spectra that arise from ensembles of conformers.

As IR spectra primarily encode vibrational information related to chemical bonds and functional groups rather than atomic identity, IR spectroscopy is primarily applied for qualitative analysis in real spectroscopic workflows, especially for identifying the presence of specific functional groups~\citep{coates2000interpretation, clayden2012organic}, which provides limited information about the full atomic composition, making it extremely difficult to determine the complete molecular structure based on IR data alone. Thus, in practice, structure elucidation is rarely attempted using IR spectra in isolation. Specifically, the molecular formula is typically established beforehand using other techniques such as elemental analysis or complementary spectroscopic data, and is then used as a known condition when interpreting IR spectra~\citep{field2013organic, Kind2007}.

Therefore, in our setting, we assume that the atom types $\mathbf{h}$ and the atom count $N$ of a molecule are known, and focus on modelling the conditional distribution over atomic coordinates $\mathbf{x}$. In other words, our problem can be formulated as $p_\theta(\mathbf{x}|S, \mathbf{h})$. We believe that this assumption reflects a realistic, standard, and experimentally grounded setting rather than an artificial constraint, and furthermore, is fully consistent with existing work on spectrum-to-structure recovery. It is common practice to use the molecular formula as an input along with a single spectrum, which is not only for the IR spectrum conditioned structure prediction~\citep{alberts2024leveraging, wu2025transformer, alberts2024unraveling}, but also for mass spectrometry conditioned molecular generation~\citep{bohde2025diffms} and NMR spectrum inverse problems~\citep{jonas2019deep}. In all of these settings, the atomic composition is treated as part of the available prior information, and the model focuses on resolving the remaining structural ambiguity given that composition.

\paragraph{Diffusion models.}
\label{sec: diff-models}
Diffusion models~\citep{sohl2015deep,ho2020denoising} are probabilistic models modelling two processes, a diffusion process and a denoising process, which can be described as the forward and reverse processes of a Markov chain with a fixed length $T$. In the forward process, the data sample $\mathbf{x}$ is mapped to a series of intermediate variables $\mathbf{x}_1\dots\mathbf{x}_T$ of the same dimensionality by gradually adding noise $\epsilon$ sampled from a standard normal distribution, 
\begin{equation}
    q(\mathbf{x}_t|\mathbf{x}_{t-1}) = \mathcal{N}(\mathbf{x}_t ; \sqrt{1-\beta_t}\mathbf{x}_{t-1}, \beta_t I),
\end{equation}
where $\beta_t$ is a hyperparameter controlling the extent of noise blending.
With sufficient steps $T$, the distribution of variable $\mathbf{x}_T$ converges to a standard normal distribution, $q(\mathbf{x}_T)\approx \mathcal{N}(0,I)$, which is also the initial state of the denoising process. The true reverse distribution $q(\mathbf{x}_{t-1} | \mathbf{x}_t)$ is approximated as normal distributions,
\begin{equation}
    p_\theta (\mathbf{x}_{t-1} | \mathbf{x}_t) = \mathcal{N}(\mathbf{x}_{t-1} ; \mu_\theta(\mathbf{x}_{t}, t), \rho_t^2 I),
\end{equation}
where $\mu_\theta$ is a neural network that computes the mean of the normal distribution, and typically the term $\rho_t$ is predetermined. 
A network $\epsilon_\theta$ predicting noise $\epsilon$ can be obtained through parameterization from $\mu_\theta$, \ie
$\mu_\theta = \frac{1}{\sqrt{1-\beta_t}}\mathbf{x}_t - \frac{\beta_t}{\sqrt{1-\alpha_t}\sqrt{1-\beta_t}}\epsilon_\theta$, where $\alpha_t = \prod_{i=1}^t1-\beta_i$. 
The model $\epsilon_\theta$ is trained to minimise a simplified variational bound, resulting in the following objective:
\begin{equation}
\mathcal{L}_{DM}=\mathbb{E}_{\mathbf{x},\epsilon\sim\mathcal{N}(0,I),t}\left[\|\epsilon-\epsilon_\theta(\mathbf{x}_t,t)\|^2\right].
\end{equation}
Latent diffusion models (LDMs)~\citep{rombach2022high} improve efficiency by performing the diffusion process in a lower-dimensional latent space. An autoencoder is first trained, where the encoder $\mathcal{E}_\phi$ maps the input data $\mathbf{x}$ to a latent representation $\mathbf{z} = \mathcal{E}_\phi(\mathbf{x})$, and the decoder $\mathcal{D}_\delta$ reconstructs the data as $\tilde{\mathbf{x}} = \mathcal{D}_\delta(\mathbf{z})$.
Diffusion is then applied in the latent space, with the denoising network $\epsilon_\theta$ training objective as:
\begin{equation}
\mathcal{L}_{LDM}=\mathbb{E}_{\mathcal{E}(\mathbf{x}),\epsilon\sim\mathcal{N}(0,I),t}\left[\|\epsilon-\epsilon_\theta(\mathbf{z}_t,t)\|^2\right].
\end{equation}
Xu \etal~\cite{xu2023geometric}~propose GEOLDM, extending latent diffusion models to molecular geometries $G = \langle \mathbf{x}, \mathbf{h} \rangle$. The encoding and decoding processes are formulated as 
$q_\phi(\mathbf{z}_\mathbf{x}, \mathbf{z}_\mathbf{h}|\mathbf{x}, \mathbf{h})= \mathcal{N}(\mathcal{\mathcal{E}_\phi}(\mathbf{x}, \mathbf{h}),\sigma_0I)$ and $p_\delta(\mathbf{x}, \mathbf{h}|\mathbf{z}_\mathbf{x}, \mathbf{z}_\mathbf{h}) = \prod_{i=1}^{N} p_\delta(x_i, h_i|\mathbf{z}_\mathbf{x}, \mathbf{z}_\mathbf{h})$ respectively.
To ensure translation invariance, the latent representation $\mathbf{z}_\mathbf{x}$ and reconstructed coordinates $\mathbf{x}$ are constrained to the subspace where the centre of mass is zero, \ie, $\sum_i \mathbf{z}_{\mathbf{x},i} = 0$ (or $\sum_i \mathbf{x}_i = 0$). This constraint is also applied to all intermediate diffusion steps.
The denoising model is then trained with the loss:
\begin{equation}
    \mathcal{L}_{LDM} = \mathbb{E}_{\mathcal{E}(G), \epsilon \sim \mathcal{N}(0,I), t} \left[ \|\epsilon - \epsilon_\theta(\mathbf{z}_{\mathbf{x},t}, \mathbf{z}_{\mathbf{h},t}, t) \|^2 \right].
\end{equation}

Following the problem definition introduced earlier, our recovery task fundamentally differs from molecular generation. It requires strong controllability rather than encouraging diversity, while the latent-diffusion formulation can improve controllability: the use of latent variables allows for better control over the generation process~\citep{rombach2022high}.
Xu \etal~\cite{xu2023geometric}~also demonstrate the higher capacity of GEOLDM for controllable generation just due to the latent modelling. Our method therefore builds on GEOLDM, which constructs a latent space consisting of both invariant scalars and equivariant tensors, ensuring strict preservation of roto-translational symmetry while has stronger controllability.

It is worth noting that in \textit{de novo} molecular generation models like GEOLDM~\citep{xu2023geometric}, both atomic coordinates $\mathbf{x}$ and atomic features $\mathbf{h}$ (\ie, atom types) are treated as variables to be generated. The model is trained to learn the joint distribution $p_\theta(\mathbf{z_x}, \mathbf{z_h})$, where the denoise network $\epsilon_{\theta}$ predicts noise $\epsilon=[\epsilon_x, \epsilon_h]$ for both the position and feature components simultaneously. During the forward diffusion process, noise is added to both $\mathbf{z}_\mathbf{x}$ and $\mathbf{z}_\mathbf{h}$, and sampling starts from a pair of fully noisy latents ($\mathbf{z}_{\mathbf{x},T}, \mathbf{z}_{\mathbf{h},T}$) from a standard Gaussian prior.

However, we aim to learn a probabilistic model $\theta$ that captures the conditional distribution $p_\theta(\mathbf{x}|S,\mathbf{h})$ where the atom types $\mathbf{h}$ and count $N$ are assumed to be known. Therefore, in our model, IR-GeoDiff, the forward process and objective are redefined to explicitly learn a conditional distribution $p_\theta(\mathbf{z}_\mathbf{x} | \mathbf{z}_\mathbf{h})$, in which 
the diffusion is applied only to the positions latent $\mathbf{z}_\mathbf{x}$ and the known atom types $\mathbf{z}_\mathbf{h}$ are not diffused. The atom types $\mathbf{z}_\mathbf{h}$ are instead fed as a fixed condition into the denoising network at every step along with the spectral features $S$, \ie $\epsilon_{\theta}(\mathbf{z}_{\mathrm{x},t}, t, S,\mathbf{z}_{\mathrm{h}})$.

\subsection{Proposed method}
\begin{figure}[ht]
    \centering
    \includegraphics[width=\linewidth]{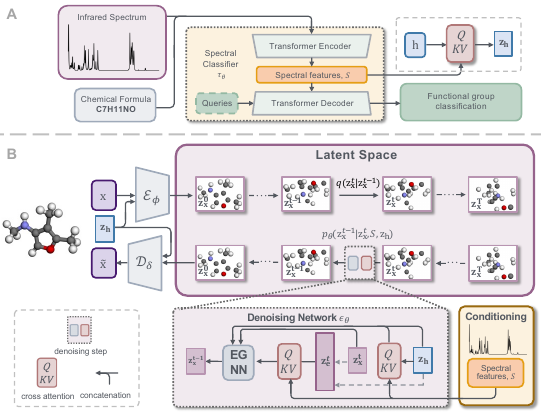}
    \caption{
    Overview of the proposed \texttt{IR-GeoDiff}. \textbf{A}: Spectral features $S$  are extracted by a Transformer-based spectral classifier $\tau_\theta$. \textbf{B}: The encoder $\mathcal{E}_\phi$ maps molecular geometries into a latent representation $\mathbf{z_x}$, which is perturbed through a forward diffusion process and denoised by an equivariant network $\epsilon_\theta$ conditioned on the spectral features which are injected into node and edge representations via cross-attention. Finally, the decoder $\mathcal{D}_\delta$ reconstructs the 3D geometry from the denoised latent representation. 
    }
    \label{fig: IR-GeoDiff}
\end{figure}

In this section, we present the proposed \texttt{IR-GeoDiff} model. We first train a Transformer-based spectral classifier~$\tau_\theta$ to extract spectral features from the input. 
Our latent diffusion model~$\epsilon_\theta$ maintains roto-translational equivariance throughout the diffusion process.
We incorporate IR spectral information into the denoising process using cross-attention mechanisms, enabling spectrum-conditioned molecular recovery. An overview is shown in Figure~\ref{fig: IR-GeoDiff}.

\paragraph{Spectral classifier and functional group representation.}
Following Wu \etal~\cite{wu2025transformer}, we adopt a patch-based embedding layer to extract local spectral features from the IR spectrum and employ a standard Transformer encoder to capture the global spectral characteristics, as shown in Figure~\ref{fig: IR-GeoDiff}A. As discussed in Section~\ref{sec: prob-define}, we incorporate the chemical formula as input.
Let $p$ be the number of spectral patches and $c$ be the length of the split chemical formula, the spectral encoder generates the spectral feature $S\in\mathbb{R}^{(p+c)\times d_s}$.
To ensure that the spectral encoder can effectively comprehend IR spectra, we implement a classifier to learn the spectral representations by categorising IR spectra according to the functional groups present within the corresponding molecules. 
We introduce functional group queries as a learnable matrix $M \in \mathbb{R}^{m \times d_\mathrm{fg}}$, where $m$ is the number of pre-defined functional groups and $d_\mathrm{fg}$ is the feature dimensionality. The queries are randomly initialised and fed into the Transformer-based decoder together with the spectral feature $S$. Through cross-attention, $M$ interacts with the spectral representations $S$ to produce functional group embeddings, which are then used for multi-label classification.
For each spectrum $S$, the multi-label $\mathrm{y}\in \{0, 1\}^m$. 
The binary cross-entropy (BCE) loss is used to solve this $m$ binary classification problem,
\begin{equation}
    \mathcal{L}_{cls} = \frac{1}{m} \sum_{k=1}^m [\mathrm{y}_k \cdot \log \tilde{\mathrm{y}}_k + (1-\mathrm{y}_k)\cdot \log (1-\tilde{\mathrm{y}}_k)],
\end{equation}
where $\tilde{\mathrm{y}}$$_k$$=\mathrm{sigmoid}(f_S^k)$, and $f_S^k$ is the classifier's output of the $k$th group.

\paragraph{Geometric auto-encoding.}
 \label{sec: ae}
As we only focus on the perturbed positions, we use the autoencoder to learn a latent space of $\mathbf{x}$. Precisely, given a geometry $G=\langle \mathbf{x}, \mathbf{h}\rangle$, we pre-process $\mathbf{h}$ to obtain latent $\mathbf{z}_\mathrm{h}$, and our encoder $\mathcal{E}_\phi$ encodes $\mathbf{x}$ along with $\mathbf{z}_\mathrm{h}$, \ie $\mathbf{z}_\mathbf{x} = \mathcal{E}_\phi(\mathbf{x}, \mathbf{z}_\mathrm{h})$. 
The decoder reconstructs the positions from the latent, \ie $\tilde{\mathbf{x}}=\mathcal{D}_\delta(\mathbf{z}_\mathrm{x}, \mathbf{z}_\mathrm{h})$, thus giving the geometry $\tilde{G}=\langle\tilde{\mathbf{x}},\mathbf{h}\rangle$.
Therefore, the encoding and decoding process can be formulated by  $q_\phi(\mathbf{z}_\mathrm{x}|\mathbf{x}, \mathbf{z}_\mathrm{h}) = \mathcal{N}(\mathcal{E}_\phi(\mathbf{x},\mathbf{z}_\mathrm{h}), \sigma_0 I)$ and $p_\delta(\mathbf{x}|\mathbf{z}_\mathrm{x}, \mathbf{z}_\mathrm{h}) = \prod_{i=1}^{N} p_\delta(x_i|\mathbf{z}_\mathrm{x}, \mathbf{z}_\mathrm{h})$ respectively.
Following Xu \etal~\cite{xu2023geometric}, the encoder $\mathcal{E}_\phi$ and decoder $\mathcal{D}_\delta$ of our autoencoder are based on equivariant graph neural networks (EGNNs)~\citep{satorras2021n} in order to incorporate equivariance into $\mathcal{E}_\phi$ and $\mathcal{D}_\delta$.
A brief overview of equivariance is provided in Appendix~\ref{sec: equi}, and architectural details of EGNNs are given in Appendix~\ref{sec: egnn}.
The autoencoder is optimised by
\begin{equation}
    \begin{split}
        &\mathcal{L}_{AE} = \mathcal{L}_{recon} + \mathcal{L}_{reg},\\
        &\mathcal{L}_{recon} = -\mathbb{E}_{q_\phi(\mathbf{z}_\mathrm{x}|\mathbf{x}, \mathbf{z}_\mathrm{h})}p_\delta(\mathbf{x}|\mathbf{z}_\mathrm{x}, \mathbf{z}_\mathrm{h}),
    \end{split}
\end{equation}
where $\mathcal{L}_{reg}$ is KL-reg~\citep{rombach2022high}, a slight Kullback-Leibler penalty of $q_\phi$ towards standard Gaussians similar to variational auto-encoders, and $\mathcal{L}_{recon}$ in practice is calculated as $L_2$ norm of the input $\mathbf{x}$ and the decoded $\tilde{\mathbf{x}}$. 

To obtain the latent atomic representations $\mathbf{z}_\mathrm{h}$, we follow the tokenisation approach analogous to that used in natural language processing. Specifically, we first construct a vocabulary of all atom types appearing in the dataset. Each atom type is treated as a discrete token and mapped to a $d_h$-dimensional embedding via a learnable embedding matrix $T \in \mathbb{R}^{n \times d_h}$, where $n$ is the vocabulary size. Given a molecule with $N$ atoms, this results in atomic features $\mathbf{z}_\mathrm{h} \in \mathbb{R}^{N \times d_h}$.
Spectral information is then introduced to atomic features $\mathbf{z}_\mathrm{h}$ by multi-head cross-attention. For attention head $i$, with $Q= W_Q^{(i)} \cdot \mathbf{z}_\mathrm{h}, K=W_K^{(i)} \cdot S, V=W_V^{(i)}\cdot S$,
\begin{equation}
    \mathrm{Attention}(Q,K,V) = \mathrm{softmax}\left(\frac{QK^T}{\sqrt{d}}\right)\cdot V,
\end{equation}
where $W_Q^{(i)} \in \mathbb{R}^{d\times d_h}$, $W_K^{(i)}, W_V^{(i)} \in \mathbb{R}^{d\times d_s}$ are learnable projection matrices.

\paragraph{IR spectra conditioned latent diffusion.}
We introduce spectral information to both node features and edge features. 
Edge features $\mathbf{z}_{e}\in\mathbb{R}^{N(N-1)\times d_e}$ are constructed exclusively from invariant quantities. For an edge between atoms $i$ and $j$, the feature $\mathbf{z}_{e_{ij}}$ is generated from 
the squared distance $\| \mathbf{z}_{\mathrm{x}_i} - \mathbf{z}_{\mathrm{x}_j}\|^2$ and the sum of invariant latent features $\mathbf{z}_{\mathrm{h}_i}$ and $\mathbf{z}_{\mathrm{h}_j}$,  \ie $\mathbf{z}_{e_{ij}}= \text{MLP}(\text{concat}(\| \mathbf{z}_{\mathrm{x}_i} - \mathbf{z}_{\mathrm{x}_j}\|^2, \mathbf{z}_{\mathrm{h}_i}+\mathbf{z}_{\mathrm{h}_j}))$. By construction, this design guarantees that edge features are invariant to both rotation and translation.

To inject spectral information into the denoising process while preserving SE(3)-equivariance, we use the invariant atomic type embeddings $\mathbf{z}_{\mathrm{h}}$ to interact with the spectral features $S$ via cross-attention. The resulting invariant representation $\mathbf{z}_{hs}$ is concatenated with $\mathbf{z}_{\mathrm{h}}$ and passed to the EGNN backbone.
We further apply cross-attention to incorporate spectral features~$S$ into the edge representations $\mathbf{z}_e$. 

We adopt EGNNs as the backbone of the denoising network $\epsilon_\theta$, which preserves both invariance and equivariance.
Conditioning on IR spectra $S$ 
, we then learn the conditional LDM via
\begin{equation}
    \mathcal{L}_{LDM}=\mathbb{E}_{\mathcal{E}(\mathbf{x}),\epsilon\sim\mathcal{N}(0,I),t}\left[\|\epsilon-\epsilon_\theta(\mathbf{z}_\mathrm{x}^t,t,{\mathbf{z}_\mathrm{h}},S)\|^2\right].
\end{equation}

\subsection{Training and sampling}
\label{sec: procedures}
\paragraph{Training.}
We pretrain the spectral classifier~$\tau_\theta$ using the classification loss $\mathcal{L}_{cls}$. During autoencoder training, the classifier~$\tau_\theta$ is jointly optimised with the autoencoder to ensure the alignment with the learned spectral features. The overall objective in this stage is $\mathcal{L}_{AE} + \mathcal{L}_{cls}$.
In the subsequent diffusion training stage, the spectral classifier is frozen to ensure a stable and consistent conditioning signal, while the autoencoder remains learnable. The model is optimized with $\mathcal{L}_{AE} + \mathcal{L}_{LDM}$.
\paragraph{Sampling.}
Given the number $N$ and the types $\mathbf{h}$ of atoms, we first sample a latent code $\mathbf{z}_\mathrm{x}^T$ from the standard normal distribution. This latent is then denoised iteratively by using $\mathbf{z}_\mathrm{x}^t$ by $\mathbf{z}_\mathrm{x}^{t-1} = \frac{1}{\sqrt{1-\beta_t}}(\mathbf{z}_\mathrm{x}^t - \frac{\beta_t}{\sqrt{1-\alpha_t}}\epsilon_\theta(\mathbf{z}_\mathrm{x}^t,t,\mathbf{z}_\mathrm{h},S))+\rho_t\epsilon$. After the denoising process, the latent $\mathbf{z}_\mathrm{x}$ is decoded into the molecular geometry via $p_\delta(\mathbf{x}|\mathbf{z}_\mathrm{x}, \mathbf{z}_\mathrm{h})$, resulting in a final geometry $G=\langle \mathbf{x}, \mathbf{h}\rangle$.
\subsection{Evaluation metrics for conditional results}
We evaluate conditioning performance from two perspectives: structural similarity and spectral similarity.
As our goal is to recover molecular structures consistent with a given IR spectrum, we expect the sampled molecules to exhibit both high structural similarity to the reference molecule and high spectral similarity to the input spectrum.

\paragraph{Structural similarity.}
Due to the theoretically one-to-one correspondence between molecular structures and their IR spectra (ignoring enantiomers), we evaluate the structural similarity between sampled molecules and the reference structures. 
After sampling geometries $\tilde{G}=\langle \mathbf{x}, \mathbf{h}\rangle$, we further convert them into a molecular graph using a valence-guided bond assignment algorithm, as detailed in Appendix~\ref{sec: xyz2mol}.
Following~Adams \etal~\cite{adams2023equivariant} and~Chen \etal~\cite{ chen2023shapeconditioned}, we measure chemical graph similarity $\mathrm{sim}_g(\tilde{G}, G)$, defined as the Tanimoto similarity between Morgan fingerprints computed using RDKit~\citep{rdkit}. 
Inspired by the top-\textit{n} accuracy metrics used in prior works~\citep{alberts2024leveraging, wu2025transformer, alberts2025setting}, we further define \textit{molecular accuracy} to evaluate the model's overall performance across the test set. For each input IR spectrum, we generate multiple molecular samples; if at least one sampled molecule exactly matches the reference structure, the test case is considered successful. 
Following previous IR-to-structure work~\citep{alberts2024leveraging, wu2025transformer, alberts2025setting}, we define a molecule as correct if its canonical SMILES string is the same as the canonical SMILES of the reference molecule. Canonical SMILES provides a unique representation of the molecular graph, which allows us to determine chemical identity in a consistent and unambiguous manner across all generated samples.

\paragraph{Spectral similarity.}
To measure spectral similarity, we adopt the Spectral Information Similarity (SIS) metric $\mathrm{SIS}(S_{G}, S)$ proposed by~McGill \etal~\cite{mcgill2021predicting}, which quantifies the similarity between the given spectrum $S$ and the spectra of the sampled molecule $S_G$. 
Further details of computing SIS are provided in Appendix~\ref{sec: sis}. In typical IR spectral analysis, chemists primarily focus on the functional group region (1,350–4,000 cm\textsuperscript{-1}), where distinct peaks corresponding to specific functional groups or chemical bonds are found. The fingerprint region (400–1,350 cm\textsuperscript{-1}) is typically used for final confirmation of molecular identity~\citep{clayden2012organic}. We further define SIS\textsuperscript{*}, which computes the SIS score restricted to the functional group region. 
It is worth noting that SIS also serves as a physically grounded measure of geometric correctness.
Since IR spectra are computed directly from the full three-dimensional geometry, any geometric or conformational discrepancy will manifest in the resulting spectrum. In other words, an inaccurate 3D geometry will lead to a mismatched spectrum and a lower SIS score. 

\subsection{Datasets} 
While datasets used for molecular generation tasks, such as the  GEOM dataset~\citep{axelrod2022geom}, contain larger and more diverse molecular structures, they do not provide corresponding IR spectra, which are crucial for our task. Conversely, datasets utilised in prior IR-to-structure studies usually include IR spectra and molecular structures as SMILES strings but lack 3D geometries. Therefore, we only use the QM9S dataset and QMe14S dataset, which offer both IR spectra and corresponding 3D molecular geometries, to train and evaluate our method.

QM9S dataset~\citep{zou2023qm9s} is derived from the QM9 dataset~\citep{ramakrishnan2014quantum} and contains approximately 130k small molecules composed of five atom types (H, C, N, O, and F), with up to nine heavy atoms. QM9S extends QM9 by providing multiple types of spectra. In particular, IR spectra are obtained by re-optimising molecular geometries and performing vibrational analysis using quantum chemical calculations with Gaussian 16~\citep{g16}. 
QMe14S dataset~\citep{yuan2025qme14s} provides 186,102 IR spectra and corresponding 3D structures for molecules that are larger in size and cover 14 different elements (H, B, C, N, O, F, Al, Si, P, S, Cl, As, Se, and Br). As most QM9S molecules are also included in QMe14S, we remove those that appear in the QM9S training set to ensure that the QMe14S test set does not overlap with the data used for pretraining. After this filtering, the QMe14S subset used in our experiments contains 53,344 molecules.

Prior to training, we perform a data screening step to remove invalid molecules. 
Following~\citep{ayadi2024unified,chen2023shapeconditioned}, we then shuffle all samples and randomly select 1,000 molecules as the test set. The remaining molecules are split into training and validation sets with a 95:5 ratio. Further details of datasets are provided in Appendix~\ref{sec: dataset}. 

\section{Results and discussion}
\label{sec: exp-setting}

In this section, we report results based on the proposed metrics that evaluate how well the sampled molecules match the input IR spectra. Structural quality metrics (e.g. validity, stability, and connectivity) that assess whether a generated geometry is physically meaningful, are reported in Appendix~\ref{sec: common_metrics}. Spectra of sampled molecules are computed using the same procedure as in QM9S. All evaluations are performed on stable molecules only, as unstable samples often undergo significant structural changes during geometry optimisation with Gaussian 16~\citep{g16}, resulting in spectra that no longer reflect the original structures. 

Due to the high computational cost of quantum chemical calculations with Gaussian 16 as detailed in Appendix~\ref{sec: gaussian-cal}, SIS and SIS\textsuperscript{*} reported in Table~\ref{tab: comparison} and Table~\ref{tab: qme14s} are computed on a subset of 200 test spectra, each with 50 sampled molecules (appr. 10,000 spectra in total).
Although a fast, deep-learning-based spectral predictor such as Chemprop-IR~\citep{mcgill2021predicting} and DetaNet~\citep{zou2023qm9s} can be utilized to replace the quantum chemistry (QC) calculation to reduce the required computation time and resources, it may entangle the generative error of IR-GeoDiff with the prediction error of the surrogate model, making it much harder to interpret whether discrepancies arise from our method or from the metric itself. Therefore, in this initial study, we chose to prioritise physical rigour and deliberately adopt the QC-based SIS as a ``gold-standard'' spectral metric, complemented by structural metrics such as sim$_g$ and molecular accuracy. 

\subsection{Recovery of molecular geometries from IR spectra}
\label{sec: recovery}

\paragraph{Baselines.}
\begin{table}[t]
  \caption{Comparison with baseline models and ablation variants models on QM9S dataset. \textit{edge ca}: cross-attention between edge features and functional group features; \textit{atom ca}: cross-attention between atomic node features and IR spectral features.}
  \label{tab: comparison}
  \centering
  \small
  \begin{tabular}{lcccccc}
    \toprule
    Method 
    & $\mathrm{sim}_g$  & $\max \mathrm{sim}_g$   
    & $\mathrm{mol\ acc}$(\%) & SIS 
    & $\max$ SIS & SIS\textsuperscript{*}\\
    \midrule
    EDM~\citep{hoogeboom2022equivariant} & 0.278 & 0.624 & 19.03 & 0.411 & 0.681 & 0.432 \\
    GEOLDM~\citep{xu2022geodiff} 
    & 0.355 & 0.779 & 44.47 & 0.464 & 0.780 & 0.494
    \\
    GFMDiff~\citep{xu2024geometric} & 0.364 & 0.778 & 44.37 & 0.482 & 0.791 & 0.517 \\
    EDM-variant & 0.501 & 0.938 & 81.07 & 0.576 & 0.879 & 0.619 \\
    GEOLDM-variant & 0.539 & 0.950 & 84.27 & 0.588 & 0.897 & 0.630 \\
    \midrule
    ours w/o edge ca&  0.630 & 0.977 & 91.90 & 0.638 & 0.929 & 0.680
    \\
    ours w/o atom ca& 0.605 & 0.977 & 92.27 & 0.635 & 0.928 & 0.680
    \\
    \textbf{ours} & \textbf{0.666} & \textbf{0.987} & \textbf{95.33}
     & \textbf{0.675} & \textbf{0.947} & \textbf{0.718} \\
    \bottomrule
  \end{tabular}
\end{table}

\begin{figure}[h]
    \centering
    \includegraphics[width=\linewidth]{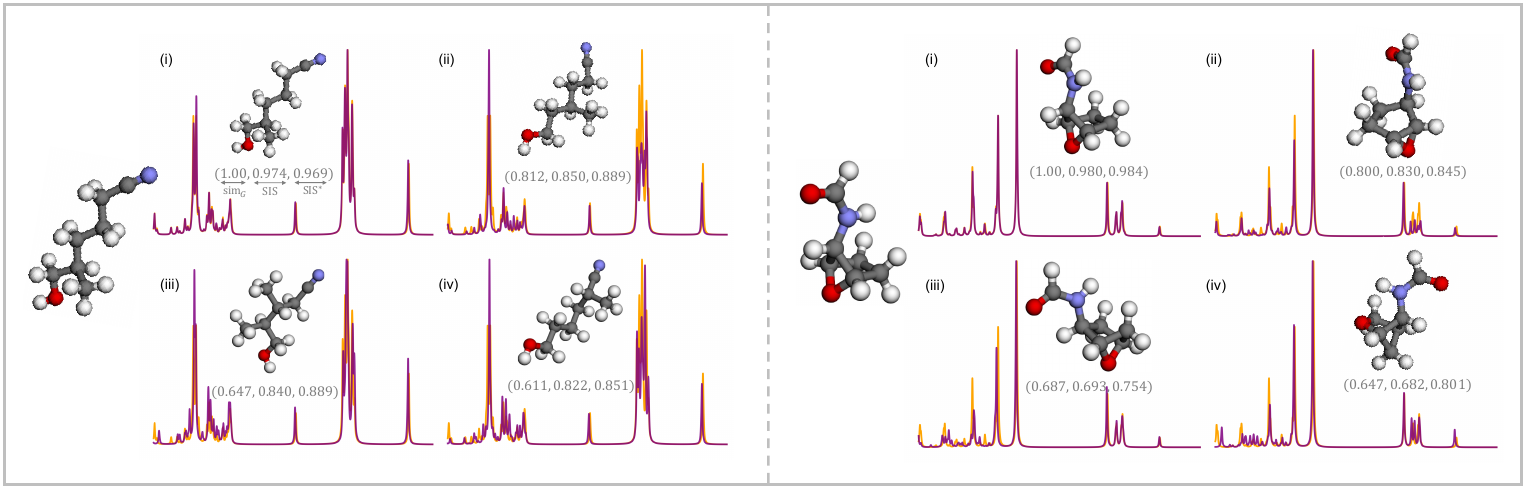}
    \caption{Examples of molecules sampled under IR spectral conditions. For spectra, \textcolor{orange}{orange} curves denote input IR spectra; \textcolor{purple}{purple} curves denote spectra of sampled molecules. For atoms, hydrogen: {white}~\hydrogendot, \textcolor{gray}{carbon: {gray}}~\atomdot{darkgray}, \textcolor{red}{oxygen: {red}~\atomdot{red}}, \textcolor{Periwinkle}{nitrogen: {blue}~\atomdot{Periwinkle}}.}
    \label{fig: sample_results}
\end{figure}

We compare IR-GeoDiff against two representative 3D molecular generative models: EDM~\citep{hoogeboom2022equivariant} and GEOLDM~\citep{xu2022geodiff}. EDM operates directly in atomic coordinate space, while GEOLDM performs generation in a learned latent space.  
In addition, we further compare IR-GeoDiff to a recent equivariant diffusion model, GFMDiff~\citep{xu2024geometric}, which incorporates both pairwise distances and triplet-wise angles in the denoising kernel to capture higher-order local geometric interactions. These multi-body patterns are closely related to functional-group-specific vibrational modes, making GFMDiff a relevant architecture to test in our IR-conditioned setting. 
We follow their conditioning mechanisms by concatenating conditioning features to the node representations.


For each test IR spectrum, we sample 50 molecules, and the results are reported in Table~\ref{tab: comparison}. Additional results with varying numbers of samples are provided in Appendix~\ref{sec: diff-sample}. 
As shown in Table~\ref{tab: comparison}, IR-GeoDiff achieves a molecular accuracy of 95\%, demonstrating its ability to accurately recover the distribution of molecular geometries consistent with the input IR spectrum. Compared to GEOLDM, our model improves SIS by 0.211 and SIS* by 0.224, with the larger gain on SIS\textsuperscript{*} highlighting its stronger performance in the functional group region, which is a more structurally informative part of the spectrum. Some sampling results are illustrated in Figure~\ref{fig: sample_results}. 

Across all metrics, our model consistently outperforms all baselines. 
A key reason for the performance gap is that baseline models are designed to generate diverse molecules, even under conditional settings, and do not constrain atom types or counts during generation as discussed in Section~\ref{sec: diff-models}. 
In contrast, our IR-GeoDiff aims to \textit{recover a precise molecular distribution defined by the given IR spectrum}. Thus, both atom types and counts are provided as part of the input to reduce the search space. For a fair comparison, we modify the baseline sampling procedures by fixing the number of atoms, but remain atom types unconstrained because their denoising networks jointly model position and feature-level noise, which does not support atom types being specified as input. 

\paragraph{Variants.} To better understand the role of atomic-type constraints, we introduce variant versions of EDM and GEOLDM that constrain atom types while keeping the original models unchanged. 
As discussed in Section~\ref{sec: diff-models}, sampling in GEOLDM starts from a pair of fully noisy latents $(\mathbf{z}^T_\mathrm{x}, \mathbf{z}^T_\mathrm{h})$ drawn from a standard Gaussian prior. If atom types are constrained, one would instead require noisy $\mathbf{z}^T_\mathrm{x}$ but clean $\mathbf{z}_\mathrm{h}$, \ie $(\mathbf{z}^T_\mathrm{x}, \mathbf{z}_\mathrm{h})$. However, the autoencoder in GEOLDM encodes a clean pair $(\mathbf{x}, \mathbf{h})$ into $(\mathbf{z}_\mathrm{x}, \mathbf{z}_\mathrm{h})$. Since $\mathbf{x}$ is unknown during sampling, the clean $\mathbf{z}_\mathrm{h}$ cannot be obtained. Thus, a new autoencoder is required to obtain $\mathbf{z}_\mathrm{h}$ independent of $\mathbf{x}$. To ensure the fairness of comparison as much as possible, we therefore adopt the same latent initialisation scheme of \texttt{IR-GeoDiff} as detailed in Section~\ref{sec: ae}.
The results of the EDM-inspired and GEOLDM-inspired variants are reported in Table~\ref{tab: comparison}. 
We observe that constraining the atom types do improve the sampling performance, supporting our problem formulation where the atomic composition is treated as given prior information for the recovery task. 
However, despite incorporating the same atomic-type constraint, both variants still underperform our full model. 
This demonstrates that the performance gain cannot be solely attributed to the constraint on atom types, but also arises from the specific architectural and modelling design of our method.

    

\paragraph{Ablation study.}

We perform ablation studies to assess the contributions of key components in our denoising network, including the cross-attention between atomic features and spectral features, and the cross-attention between edge features and spectral features.
As shown in Table~\ref{tab: comparison}, removing either module leads to a performance drop across both spectral and structural metrics. This suggests that both modules contribute to reducing ambiguity in the inverse mapping from spectrum to structure, allowing the model to better recover the correct molecular distribution.

\begin{table}[t]
  \caption{Comparison with EDM and GEOLDM variant models on QMe14S dataset.
  }
  \label{tab: qme14s}
  \centering
  \small
  \begin{tabular}{lcccccc}
    \toprule
    Method 
    & $\mathrm{sim}_g$  & $\max \mathrm{sim}_g$   
    & $\mathrm{mol\ acc}$(\%) & SIS 
    & $\max$ SIS & SIS\textsuperscript{*}\\
    \midrule
    
    EDM-variant & 0.354 & 0.820 & 64.17 & 0.329 & 0.581 & 0.405 
    \\
    GEOLDM-variant & 0.551 & 0.924 & 82.83 & 0.409 & 0.641 & 0.532
    \\
    ours & \textbf{0.657} & \textbf{0.958} & \textbf{90.70} & \textbf{0.464} & \textbf{0.659} & \textbf{0.607} \\
    
    \bottomrule
  \end{tabular}
\end{table}

\paragraph{QMe14S subset.} We further fine-tune EDM-variant, GEOLDM-variant and our model on the QMe14S subset, which contains larger and chemically more diverse molecules than QM9S. As shown in Table~\ref{tab: qme14s}, our model can achieve strong performance even on these more complex molecules, demonstrating its ability to generalise beyond the QM9S domain. 

\subsection{Understanding spectral-structural relationships via attention visualisation}
\label{sec: attn}

\begin{figure}[ht]
    \centering
    \includegraphics[width=1.0\linewidth]{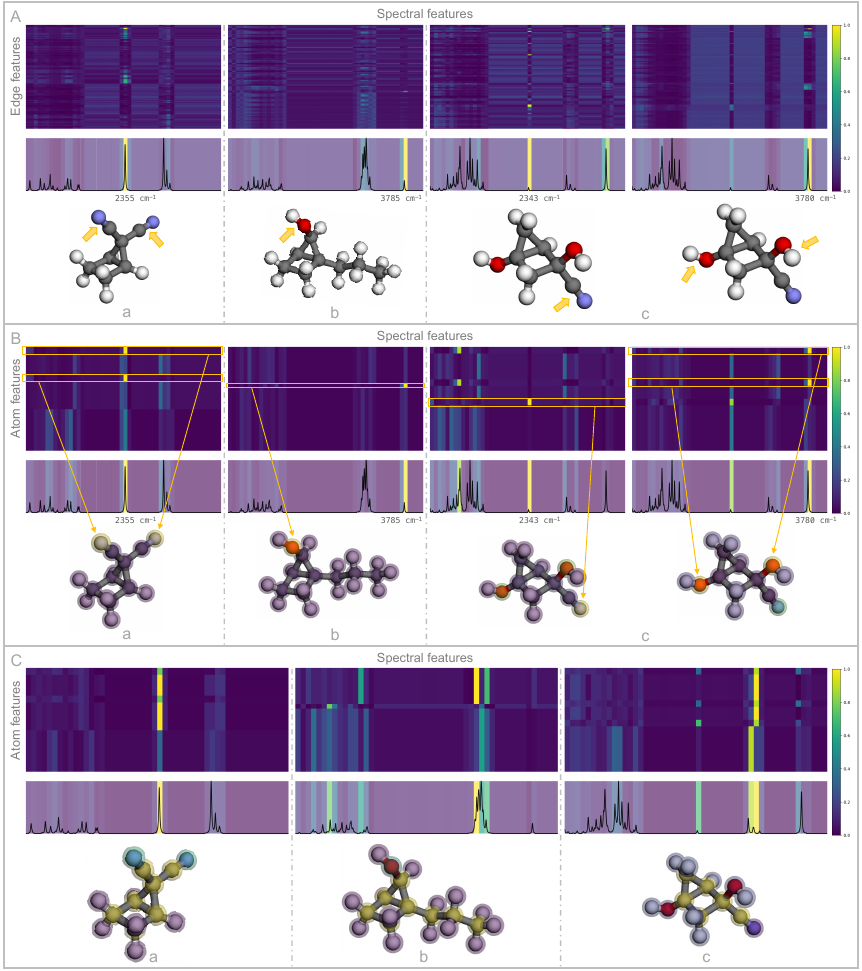}
    \caption{Visualisation of cross-attention between spectral features and molecular structure representations. 
    \textbf{A}: Spectral–edge cross-attention highlighting functional-group-specific vibrational signatures.
    \textbf{B}: Spectral–atom cross-attention revealing associations between characteristic peaks and non-carbon heavy atoms.
    \textbf{C}: Spectral–atom cross-attention emphasising the carbon backbone.}
    \label{fig: attn_map}
\end{figure}

\paragraph{Visualisation of cross-attention maps between spectral and edge features.}
As shown in Figure~\ref{fig: attn_map}A, we visualise the normalised attention maps from the cross-attention layers between spectral patch features and edge features. 
For each spectral feature, we aggregate attention scores by taking the maximum value across all edges, thereby identifying the spectral regions that receive the strongest model focus. Notably, the peaks with the highest attention weights align with characteristic functional group frequencies.

There are two representative vibrational modes in Figure~\ref{fig: attn_map}: the stretching vibration of the carbon–nitrogen triple bond (\ch{C+N}) and the stretching vibration of the \ch{O-H} bond in a hydroxy group. These correspond to absorption peaks typically observed around 2,250 and 3,600  cm\textsuperscript{-1} respectively~\citep{clayden2012organic}. As all IR spectra are computed using Gaussian 16~\citep{g16}, we further examine the vibrational mode outputs to confirm that the attended peaks indeed arise from the expected functional groups (see Appendix~\ref{sec: gaussian-output}).

For molecules \textit{a} and \textit{b}, which each contain only a single type of functional group, the model correctly identifies the corresponding characteristic spectral peak. Furthermore, molecule \textit{c} contains two different functional groups, and we observe that different cross-attention layers attend to different spectral regions associated with each group. This layer-wise specialisation indicates that its ability to disentangle and localise multiple spectral signatures within a single molecule.

\paragraph{Visualisation of cross-attention maps between spectral and atom features.}
We further visualise the normalised cross-attention maps between spectral patch features and atom features to analyse how spectral information is grounded at the atomic level. For visualisation clarity, the atomic colour intensities correspond to attention scores that are re-normalised across atoms for each molecule, highlighting the relative importance of different atoms under a given spectral context.
As shown in Figure~\ref{fig: attn_map}B, the model consistently assigns high attention weights between characteristic spectral peaks and the corresponding non-carbon heavy atoms (e.g., O and N) across all three molecules. This suggests that the model learns chemically meaningful associations between localised spectral signatures and the atoms directly responsible for the underlying vibrational modes.

In addition, we observe that the model pays broadly distributed attention over carbon atoms in certain spectral–atomic cross-attention layers, as illustrated in Figure~\ref{fig: attn_map}C. 
This behaviour suggests that, beyond identifying specific functional groups, the model also encodes information related to the global carbon framework during structure recovery. 
Furthermore, the model can sometimes focus strongly on hydrogen atoms, as shown in the Appendix~\ref{sec: ca-hydrogen}. These observations indicate that different cross-attention layers may preferentially attend to different types of atoms. 
Rather than following a fixed atom-specific bias, the model appears to adaptively allocate attention across atom types according to the spectral evidence.

\subsection{Analysis on exceptions}
\label{sec: limitation}

\begin{figure}[ht]
    \centering
    \includegraphics[width=1.0\linewidth]{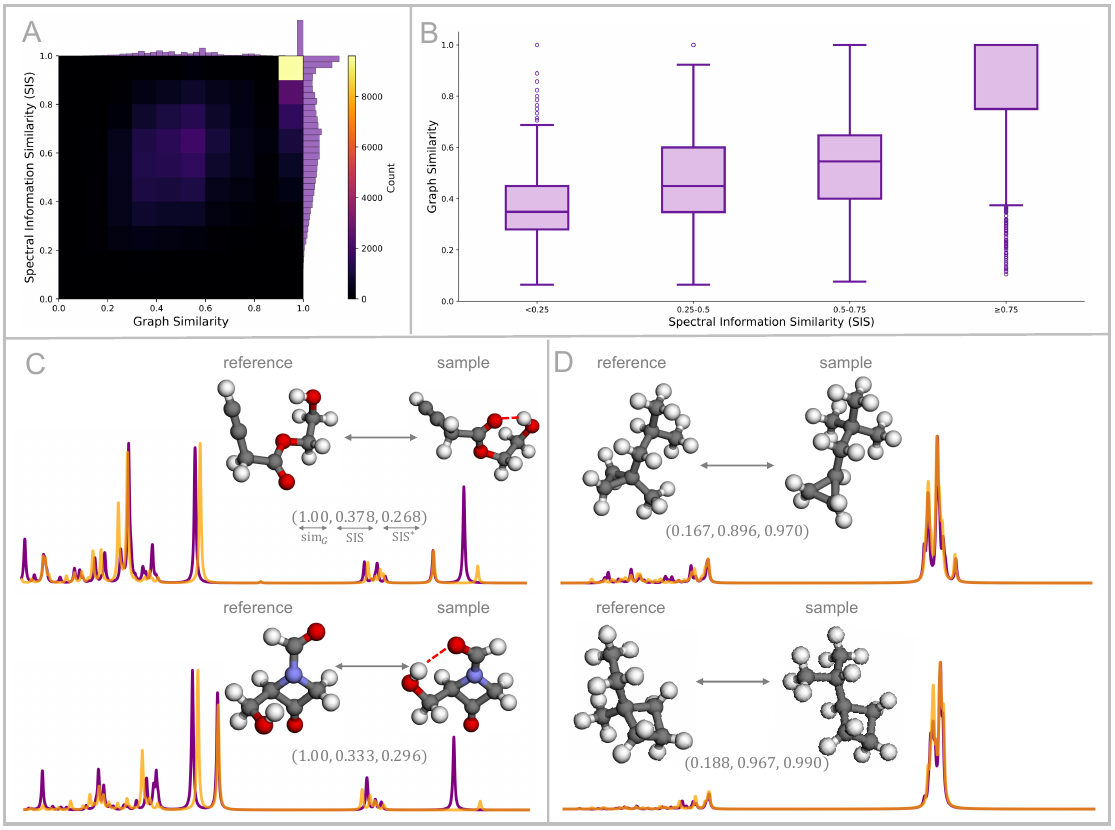}
    \caption{\textbf{A:} 2D histogram showing the joint distribution of $\mathrm{sim}_g$ and SIS, computed over 1,000 test spectra, with 50 sampled molecules per spectrum.  \textbf{B:} Box plots of $\mathrm{sim}_g$ grouped by SIS ranges. \textbf{C:} Examples with high $\mathrm{sim}_g$ but low SIS. \textbf{D:} Examples of molecules with high SIS but low $\mathrm{sim}_g$.\\  For spectra, \textcolor{orange}{orange} curves denote input IR spectra; \textcolor{purple}{purple} curves denote spectra of sampled molecules. For atoms, hydrogen: {white}~\hydrogendot, \textcolor{gray}{carbon: {gray}}~\atomdot{darkgray}, \textcolor{red}{oxygen: {red}~\atomdot{red}}, \textcolor{Periwinkle}{nitrogen: {blue}~\atomdot{Periwinkle}}.}
    \label{fig: limitation}
\end{figure}

Figure~\ref{fig: limitation}A presents the joint distribution of graph similarity $\mathrm{sim}_g$ and SIS across 1,000 test spectra, with 50 sampled molecules per spectrum.
Although there are samples lying in the mid-range for both metrics, a sharp density peak appears near the top-right corner, indicating a cluster of molecules that are near-exact matches to the reference.
As shown in Figure~\ref{fig: sample_results}, both metrics are highly sensitive to molecular configuration: even minor changes in atomic connectivity can lead to large drops in both spectral and structural similarity, making perfect matches rare and distinct.

In practical applications, chemists typically identify compounds by visually comparing the IR spectra of unknown samples with those of known references. Motivated by this, we analyse the distribution of graph similarity across different SIS ranges, as shown in Figure~\ref{fig: limitation}B. While graph similarity generally increases with SIS, the correlation between the two is moderate. This observation underscores the importance of evaluating both structural and spectral similarity simultaneously, as neither metric alone is sufficient to fully characterise the correspondence between sampled and reference molecules. Notably, there exists a non-negligible number of cases where molecules exhibit high graph similarity but low SIS, and vice versa.

\paragraph{High graph similarity but low SIS.}
As shown in Figure~\ref{fig: limitation}C, samples with high $\mathrm{sim}_g$ but low SIS, primarily caused by conformational changes, especially some of which result in the formation of intramolecular hydrogen bonds. Although hydrogen bonds are not conventional covalent bonds, they represent relatively strong electrostatic interactions, typically between hydrogen atoms and electronegative atoms such as N, O, or F atoms. These interactions can shift the vibrational frequencies of associated functional groups, leading to notable discrepancies in the IR spectra despite high structural similarity~\citep{clayden2012organic, Freedman1961}. This suggests that our model currently still has limited control over molecular conformation, and it also emphasises the importance of interpreting IR spectra within the full three-dimensional geometric context.

\paragraph{High SIS but low graph similarity.}
Samples with low $\mathrm{sim}_g$ but high SIS are shown in Figure~\ref{fig: limitation}D. This typically arises from mismatches in the molecular scaffolds, particularly when molecules lack distinctive functional groups and consist of carbon and hydrogen atoms. In such cases, the IR spectral signals reflecting differences in carbon backbone topology are often subtle and difficult to interpret, highlighting the limited ability of IR spectroscopy to resolve differences in molecular skeletons. Future work could incorporate additional spectral modalities, particularly nuclear magnetic resonance (NMR), which offers complementary information. For instance, \textsuperscript{1}H-NMR and \textsuperscript{13}C-NMR spectra are highly informative about molecular backbones and can also reveal conformational details~\citep{vakili2012conformation,tormena2016conformational}. While our model already achieves strong performance using IR spectra alone, incorporating NMR could further improve accuracy and impose stronger constraints on 3D molecular recovery.


\section{Conclusion}
In this paper, we introduce a new task of recovering the distribution of three-dimensional molecular geometries from a given infrared (IR) spectrum. To this end, we propose \texttt{IR-GeoDiff}, 
the first model that directly generates 3D molecular structures from IR spectra, together with a set of evaluation metrics measuring both structural similarity and spectral similarity. Our experiments show that IR-GeoDiff effectively captures the underlying distribution. In addition, qualitative interpretability analysis based on cross-attention visualisation indicates that the model is able to focus specially on characteristic functional group regions in IR spectra, consistent with common chemical interpretation practices.

By examining cases with high graph similarity but low SIS, and vice versa, we observe that the current model still has limited control over molecular conformations. Moreover, IR spectra contain intrinsic ambiguity for distinguishing certain molecular scaffolds. This motivates the use of additional spectral modalities, such as NMR, to provide complementary structural information and stronger constraints on 3D recovery in future work.

\section*{Acknowledgements}
This work was financially supported by the Innovation Program for Quantum Science and Technology (2021ZD0303303), the CAS Project for Young Scientists in Basic Research (YSBR-005),  the National Natural Science Foundation of China (22025304, 22033007).
The computations described in this paper were performed using the University of Birmingham's BlueBEAR HPC service, which provides a High Performance Computing service to the University's research community.
\appendix
\setcounter{figure}{0}
\setcounter{table}{0}
\renewcommand{\thetable}{A\arabic{table}}
\renewcommand{\thefigure}{A\arabic{figure}}
\startcontents[appendix]
\section*{Appendix}
\section*{Content}
\vspace{-3mm}
In this appendix, we provide additional information to support the main text. Specifically:
\begin{itemize}
    \item In Section~\ref{sec: equi}, we provide a brief overview of the concept of equivariance, which is essential for modelling molecular geometries.
    \item Section~\ref{sec: addi_details} presents the architecture details of our model, with the Equivariant Graph Neural Networks (EGNNs) and the implementation details of each component.
    \item In Section~\ref{sec: gaussian-cal}, we describe the procedure used to compute IR spectra for sampled molecules using the Gaussian 16 package.
    \item Section~\ref{sec: dataset} introduces our method for converting 3D coordinates into molecular graph, and we use it to screen and filter mismatched data in the QM9S dataset and also present the molecular size distributions of the datasets.
    \item In Section~\ref{sec: addi-results}, additional experimental results are reported, including the performance of our spectral classifier, the results of IR-GeoDiff across multiple sampling runs and different sampling counts, and standard structure quality metrics (validity, stability, connectivity), and spectral–atom cross-attention emphasising hydrogen atoms.
    \item Section~\ref{sec: gaussian-output} provides sample outputs from vibrational analysis conducted by using Gaussian 16 package.
\end{itemize}

\section{Equivariance}
\label{sec: equi}
For geometric systems like molecules, it is essential to ensure 
equivariance, as directional features such as atomic forces should transform consistently with changes in molecular coordinates~\citep{thomas2018tensorfieldnetworksrotation, weiler20183d, fuchs2020se, batzner20223}.
Let $T_g: X\rightarrow X$ and $S_g: Y\rightarrow Y$ be two sets of transformation on spaces $X$ and $Y$ respectively for an abstract group element $g\in G$. A function $\phi: X\rightarrow Y$  is equivariant to $g$ if 
\begin{equation}
    \phi(T_g(\mathbf{x})) = S_g(\phi(\mathbf{x})).
\end{equation}

For molecular geometries, denote $\mathbf{x} = (\mathrm{x}_1, \dots, \mathrm{x}_N)\in\mathbb{R}^{N\times d}$ as an input $N$ point clouds embedded in a $d$-dimensional space, and $\mathbf{y} = \phi(\mathbf{x})$ as the transformed set of point clouds with a non-linear function $\phi(\cdot)$.
In this work, we adopt the SE(3)-equivariant latent space formulation proposed by~\citep{xu2023geometric}, where both the theoretical definition and practical implementation of equivariance are derived. Specifically, 
we consider the Special Euclidean group SE(3), \ie the group of rotation and translation in 3D space, where transformations $T_g$ and $S_g$ can be represented by a translation $t$ and an orthogonal matrix rotation $R$.

For translation equivariance, translating the input by $g\in \mathbb{R}^d$ results in an equivalent translation of the output, $\phi(\mathrm{x}+g)=\phi(\mathrm{x})+g$, where $\mathrm{x}+g$ denotes the translated point cloud $(\mathrm{x}_1+g, \dots, \mathrm{x}_N+g)$. For rotation equivariance, rotating the input by an orthogonal matrix $Q\in \mathbb{R}^{d\rightarrow d}$ yields a correspondingly rotated output $\phi(Q\mathrm{x})=Q\phi(\mathrm{x})$ where $Q\mathrm{x} = (Q\mathrm{x}_1, \dots, Q\mathrm{x}_N)$.

\section{Additional Details on Experiments}
\label{sec: addi_details}

\subsection{E(n)-equivariant graph neural networks (EGNNs)}
\label{sec: egnn}
Neural networks used for the autoencoder and the backbone of our denoise network are implemented using EGNNs~\citep{satorras2021n}. Denote a molecular graph $\mathcal{G} = (\mathcal{V}, \mathcal{E})$ with nodes $v_i \in \mathcal{V}$ and edges $e_{ij} \in \mathcal{E}$. 
Each node $v_i$ is associated with a node embeddings $\mathbf{h}_i \in \mathbb{R}^{d_h}$ and a coordinate $\mathbf{x}_i \in \mathbb{R}^{d_\mathrm{x}}$ are considered.
EGNNs are composed of Equivariant Graph Convolutional Layers (EGCLs). At layer $l$, the features and coordinates are updated as  $\mathbf{h}^{l+1}, \mathbf{x}^{l+1} = EGCL[\mathbf{h}^{l}, \mathbf{x}^l, \mathcal{E}]$ with following update rules:
\begin{align}
    \mathbf{m}_{ij} &= \phi_e\left(\mathbf{h}_i^l, \mathbf{h}_j^l, d_{ij}^2, a_{ij}\right),  \\
    \mathbf{h}_i^{l+1} &= \phi_h(\mathbf{h}_i^l, \sum_{j\ne i} \tilde{e}_{ij}\mathbf{m}_{ij}), \\
    \mathbf{x}_i^{l+1} &= \mathbf{x}_i^{l} + \sum_{j\ne i}\dfrac{\mathbf{x}_i^l-\mathbf{x}_j^l}{d_{ij}+1}\phi_x(\mathbf{h}_i^l, \mathbf{h}_j^l, d_{ij}^2, a_{ij})
\end{align}
where $d_{ij}^2 = \|\mathbf{x}_i^l-\mathbf{x}_j^l\|^2$ is the squared pairwise distance and $a_{ij}$ is the edge attribute between atoms $v_i$ and $v_j$.
$\tilde{e}_{ij}=\phi_{inf}(\mathbf{m}_{ij})$ serves as the attention weights to reweight messages passed from different edges.
Following prior work~\citep{satorras2021n,hoogeboom2022equivariant,xu2023geometric}, $\mathbf{x}_i^l-\mathbf{x}_j^l$ is normalized by $d_{ij}+1$. 
$\phi_e, \phi_h, \phi_x$ and $\phi_{inf}$ are implemented as Multi Layer Perceptrons (MLPs).
EGNNs maintain equivariance to both rotations and translations of the input coordinates $\mathbf{x}_i$, and are permutation-equivariant with respect to node orderings, consistent with standard graph neural networks (GNNs).

\subsection{Model architecture details}
\label{sec: model-details}
\paragraph{Spectral classifier $\tau_\theta$.}
The spectral classifier is implemented as a Transformer~\citep{vaswani2017attention} with 4 encoder and 4 decoder layers when trained on the QM9S dataset, and 6 encoder and 6 decoder layers when trained on the QMe14S dataset. 
Both encoder and decoder use 8-head multi-head attention with a hidden dimensionality of 512. Following~Wu \etal~\cite{wu2025transformer}, we employ a patch-based spectral embedding layer to process the IR spectra: 3,200 points are uniformly sampled with a patch size of 64, resulting in 50 spectral patches.

\paragraph{Autoencoder.}
The encoder $\mathcal{E}_\phi$ and decoder $\mathcal{D}_\delta$ are implemented using EGNNs~\citep{satorras2021n}, as described in Appendix~\ref{sec: egnn}. Following the design in GEOLDM~\citep{xu2023geometric}, we use a 1-layer EGNN for the encoder and a 9-layer EGNN with 256 hidden dimensions for the decoder. The dimensionality of the latent atom embedding $\mathbf{z}_\mathrm{h}$ is 16, and the dimensionality of the coordinate latent $\mathbf{z}_\mathrm{x}$ is 3.

\paragraph{Denoising network $\epsilon_\theta$.}
The denoising network is also implemented as a 9-layer EGNN with 256 hidden dimensions. Spectra-to-node and spectra-to-edge cross-attention modules use 4 layers each. The dimensionality of the edge features is set to 16.

All models use SiLU activation functions and are implemented using the PyTorch framework~\citep{paszke2019pytorch}.

\subsection{Details of training and sampling}
\label{sec: exp-details}
All modules are trained until convergence using the AdamW optimizer~\citep{adamw}. We follow the learning rate schedule proposed by~Vaswani \etal~\cite{vaswani2017attention}, where the learning rate is defined as:
\begin{equation}
\text{rate} = 512^{-0.5} \cdot \min(\text{step}^{-0.5}, \text{step} \cdot \text{warmup\_step}^{-1.5}),
\end{equation}
with 3,000 warm-up steps.

The spectral classifier is pretrained for 100 epochs with a base learning rate of 0.8 and a batch size of 256, requiring approximately 3.5 hours. The autoencoder and diffusion model are trained with a base learning rate of 0.1 and 0.2 respectively. The autoencoder is trained for 300 epochs with a batch size of 128 (about 20 hours), and the diffusion model is trained for 1,000 epochs with a batch size of 64 (about 163 hours). Sampling 50 molecules for each of the 1,000 test spectra takes approximately 27 hours with a batch size of 128.
For the QMe14S dataset, we additionally train a spectral classifier on the combined training sets of QM9S and QMe14S for 100 epochs with a base learning rate of 0.8 and a batch size of 256, which takes about 6 hours. The autoencoder and denoising network pretrained on QM9S are then finetuned on the QMe14S subset with a base learning rate of 0.1 and 0.2 respectively. The autoencoder is finetuned for 150 epochs with a batch size of 64 (about 6 hours), and the diffusion model is finetuned for 250 epochs with a batch size of 16 (about 33 hours). Sampling 50 molecules for each of the 1,000 test spectra in QMe14S requires approximately 40 hours with a batch size of 320.

For the results of EDM, GEOLDM and GFMDiff reported in Table~\ref{tab: comparison}, spectral features are obtained from our pretrained spectral classifier. To make the features compatible with their architectures, we use an MLP to project the spectral features $S\in \mathbb{R}^{(p+c)\times d_s}$ to $\mathbb{R}^{1 \times (p+c)}$, allowing it to be concatenated with atomic features.
All models are trained until convergence. Specifically, EDM and GFMDiff are trained for 1,000 epochs, while GEOLDM is fine-tuned for 500 epochs starting from the official checkpoint provided by the authors. 
Both EDM-inspired variant and GEOLDM-inspired variant are trained for  1,000 epochs on the QM9S dataset. For the QMe14S dataset, the EDM-inspired variant is finetuned for 500 epochs, while the GEOLDM-inspired variant directly adopt our auto-encoder trained on the QMe14S dataset and further is finetuned for 250 epochs.

Experiments for training and sampling of GFMDiff are are conducted on three NVIDIA A100 GPUs and all other experiments are conducted on a single NVIDIA A100 GPU.

\section{Computation of IR Spectra}
\label{sec: gaussian-cal}
We use the Gaussian 16 package~\citep{g16} to re-optimise the molecular geometries and perform vibrational analysis of the sampled molecules at the B3LYP/def2-TZVP level of theory, consistent with the QM9S dataset~\citep{zou2023qm9s}. The same method is also used to broaden the resulting IR spectra. After applying a scaling factor of 0.965 to the computed frequencies, the IR spectra are generated using Lorentzian broadening:
\begin{equation}
    Lo(x) = \dfrac{F}{2\pi}\times\dfrac{1}{(x-x_n)^2+0.25\times F^2}\times y_n
\end{equation}
where $F$ represents the half-width of the peak, set to 15~cm\textsuperscript{-1} for the infrared spectra. $x_n$ and $y_n$ represents the calculated wavenumber and the IR intensity of the $n^{th}$ vibration mode respectively, and $x$ is any wavenumber of the IR spectra.

\begin{figure}[h]
    \centering
    \includegraphics[width=0.8\linewidth]{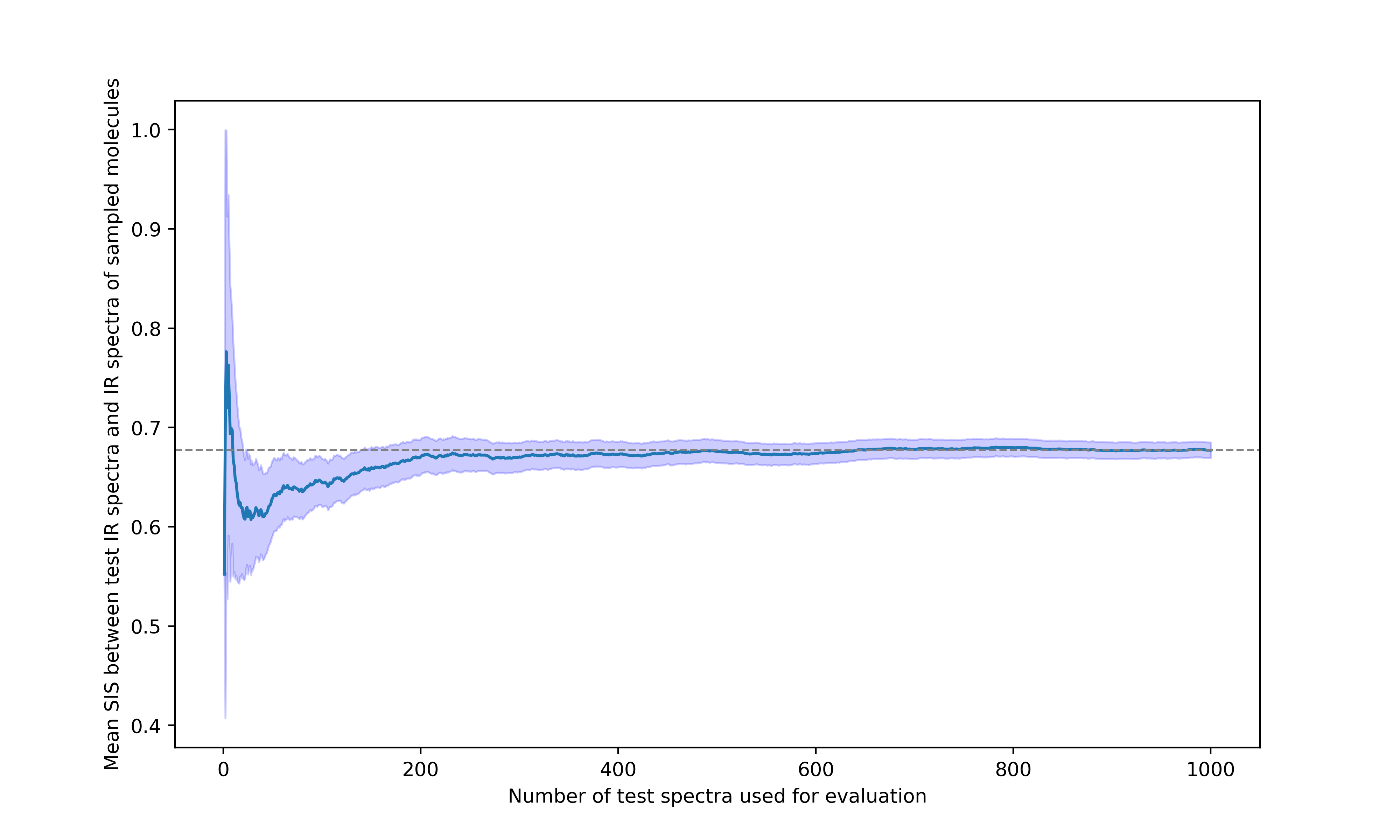}
    \caption{Convergence of the mean Spectral Information Similarity (SIS) as the number of test IR spectra increases. The shaded region indicates the 95\% confidence interval of the mean SIS. The curve stabilises after approximately 200 spectra, supporting the use of this subset for efficient and reliable evaluation. The result is based on one full sampling run (50,000 molecules in total).}
    \label{fig: meanSIS-molnum}
\end{figure}

All quantum chemical calculations were performed using Gaussian 16~\citep{g16}, with 32 GB of allocated memory (\%Mem=32GB) and 16 shared CPU cores (\%NProcShared=16). The maximum allowed computation time per job was set to 1 hour. On average, evaluating the IR spectrum of one sampled molecule takes approximately 10 minutes.
As each sampling round generates 50 molecules for each of the 1,000 test IR spectra (\ie 50,000 molecules per round), computing IR spectra for all samples is computationally expensive. Therefore, we first performed a one-time full-spectrum evaluation over all samples to assess the stability of the average SIS score on the QM9S test set.
As shown in Figure~\ref{fig: meanSIS-molnum}, the mean SIS value stabilises once the number of evaluated test spectra exceeds 200. Based on this observation, all subsequent spectral evaluations in our experiments were conducted using a subset of 200 test spectra.

\section{Dataset Preprocessing and Distribution}
\label{sec: dataset}
In this work, we explicitly include hydrogen atoms when reconstructing molecular structures, as vibrations involving hydrogens contribute significantly to IR spectral features. Since we focus exclusively on neutral molecules without formal charges or radicals, chemical bonds and their orders are inferred based on atomic valence and connectivity rules.

Although the latest version of RDKit~\citep{rdkit} provides an internal function, \texttt{xyz2mol}, for converting 3D atomic coordinates into RDKit Mol objects, it is also designed to support charged species. As a result, it occasionally fails to produce valid molecular graphs, even for geometry-optimised structures computed using Gaussian 16.

\subsection{Bond assignment procedure}
\label{sec: xyz2mol}
We assign chemical bonds and bond orders step-by-step based on atomic distances and valence constraints. Our method supports complete neutral molecules composed of atoms from the set: H, B, C, N, O, F, Si, P, S, Cl, As, Se, Br.
\begin{enumerate}
    \item \textbf{Construct initial adjacency matrix.} We construct an initial adjacency matrix by assigning a bond between atoms $i$ and $j$ if their distance $d_{ij}$ is less than a threshold $L_{ij} + \delta$, where $L_{ij}$ is the reference single bond length for that atom pair~\citep{hoogeboom2022equivariant}, and the distance tolerance is $\delta = 40$ pm for QM9S dataset and $\delta = 30$ pm for QMe14S. This yields a basic single-bond connectivity graph, which is then converted into an RDKit Mol object.
    \item \textbf{Update terminal atoms.} After constructing the initial adjacency matrix using single bond thresholds, we identify saturated atoms as those whose valence is fully occupied by current connections. This initially includes hydrogen and halogen atoms, which form only one bond. Starting from the molecular periphery, we iteratively mark all saturated atoms and then identify terminal atoms based on their local environment. A carbon, nitrogen, or oxygen atom is considered terminal if it is connected to only one neighbour that is not yet saturated. For sulfur atoms, we treat them as terminal when they are involved in only one bond. Once all terminal atoms are identified, we increase bond orders along their existing bonds until their valences are satisfied.
    \item \textbf{Update aromatic rings.} We identify all independent rings in the molecule by using RDKit and check whether they satisfy Hückel’s rule (\ie containing $4n+2$ $\pi$ electrons). For rings deemed aromatic, we assign appropriate bond orders.
    \item \textbf{Update C and N atoms.} Carbon and nitrogen are capable of forming multiple bonds, including triple bonds, and thus require additional handling. We prioritise atoms with higher unsaturation, and for each, we first attempt to increase bond orders with unsaturated neighbours. If all neighbours are saturated, we consider bonding to atoms such as sulfur or phosphorus if their valence allows. When multiple candidate bonds are equally unsaturated, we prefer shorter bonds, as higher bond orders generally correspond to shorter bond lengths.
    \item \textbf{Iterative bond order refinement.} We iteratively refine bond orders for atoms whose current valence does not correspond to any chemically valid state (e.g., carbon with five bonds, or oxygen with one). At each step, we identify such atoms and increase the bond order with neighbouring atoms that can legally accept more bonds. The process continues until all atoms satisfy one of their allowed valence states, or no further updates are possible.
    \item \textbf{Handle failures by bond removal.} If the above steps fail to produce a chemically valid molecule (e.g., incorrect charges or invalid valences), we remove the bond with the largest deviation from its reference length (if the deviation exceeds 15 pm) and restart the bond assignment process. This serves as a fallback to escape from invalid local minima.
    \item \textbf{Final validation.}
    We ensure that the final molecule is charge neutral (unless explicitly allowed) and that all valence constraints are satisfied. If no valid structure can be found within 10 iterations, the current structure is returned with a warning.
\end{enumerate}

\subsection{Screening out invalid geometries}
\label{sec: screen}

\begin{figure}[h]
    \centering
    \includegraphics[width=0.7\linewidth]{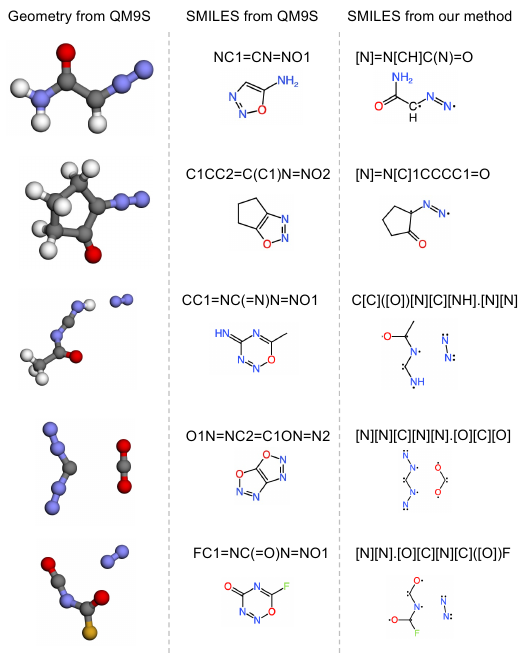}
    \caption{Examples of molecular geometries (hydrogen: {white}~\hydrogendot, \textcolor{gray}{carbon: {gray}}~\atomdot{darkgray}, \textcolor{red}{oxygen: {red}~\atomdot{red}}, \textcolor{Periwinkle}{nitrogen: {blue}~\atomdot{Periwinkle}}, \textcolor{YellowOrange}{fluorine: yellow}~\atomdot{YellowOrange}) from the QM9S dataset that are inconsistent with their provided SMILES strings. The \textbf{left} column shows 3D geometries, the \textbf{middle} column displays the corresponding SMILES from QM9S, and the \textbf{right} column shows the SMILES reconstructed by our method based on the 3D coordinates. All SMILES reconstructed by our method are consistent with the corresponding molecular geometries. (Molecular graphs converted from SMILES representations are provided, below each SMILES, for easier comparison.)}
    \label{fig: qm9s_bad}
\end{figure}

We apply our bond reconstruction method described in Section~\ref{sec: xyz2mol} to the molecular geometries provided in the QM9S dataset and identify 603 molecules with inconsistencies between the 3D coordinates and the corresponding SMILES strings. Examples are shown in Figure~\ref{fig: qm9s_bad}. 
Similarly, in the QMe14S dataset, we detect 1,873 molecules with such inconsistencies. In all experiments, we remove these invalid molecules from the datasets before training.

\newpage
\subsection{Dataset Distribution}
\label{sec: dataset-distribution}

\begin{figure}[h]
    \centering
    \includegraphics[width=1.0\linewidth]{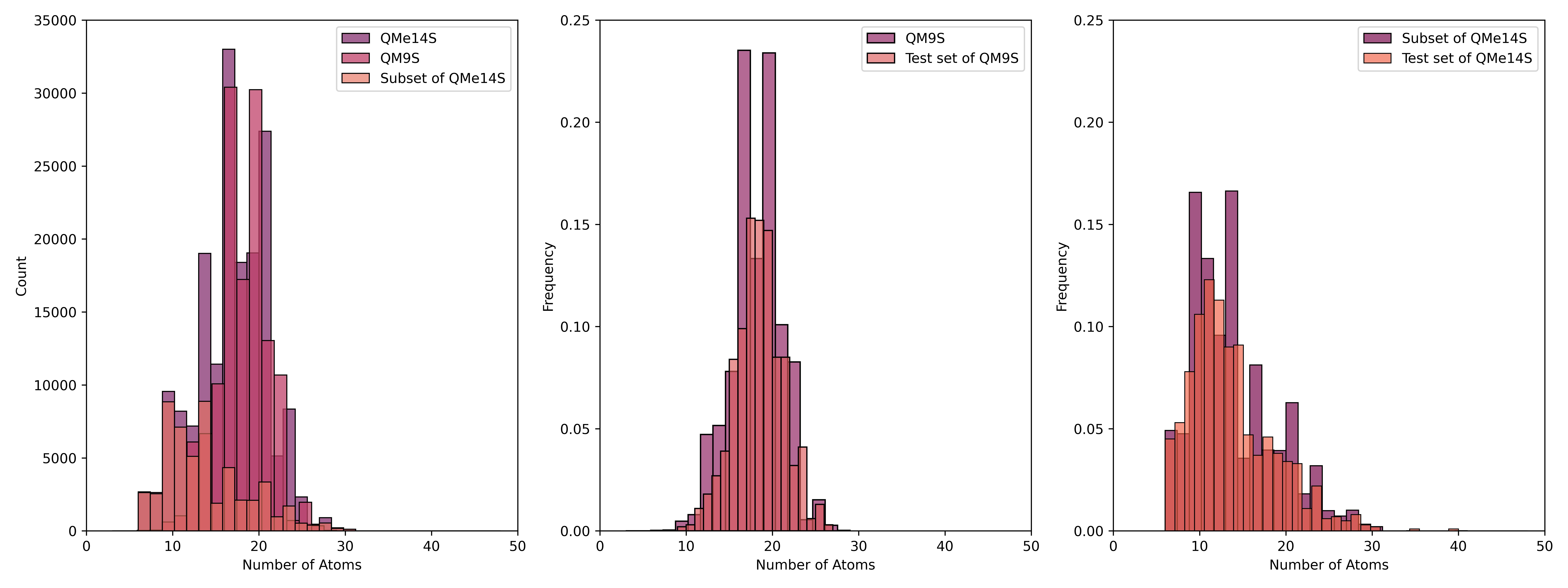}
    \caption{Distribution of molecular sizes (number of atoms). \textbf{Left}: QM9S, QMe14S, and the QMe14S subset. \textbf{Middle}: QM9S vs. QM9S test set. \textbf{Right}: QMe14S subset vs. QMe14S test set.}
    \label{fig: dataset-distribution}
\end{figure}

Figure~\ref{fig: dataset-distribution} shows the distributions of molecular sizes, measured by the number of atoms. We observe that the test sets of both QM9S and the QMe14S subset exhibit distributions that are statistically similar to their corresponding full datasets, indicating that the randomly split test sets are representative.

\section{Additional Experimental Results}
\label{sec: addi-results}
\subsection{Labelling of functional groups}
\label{sec: fg}
We use SMARTS strings to identify function groups within molecules, and define 20 types of common functional groups in molecules of QM9S dataset, as shown in Table~\ref{tab: fg-smarts}. 
For the spectral classifier trained on QMe14S dataset, 26 types of common functional groups are pre-defined as shown in Table~\ref{tab: fg-smarts-qme14s}.

\begin{table}[h]
  \caption{20 SMARTS representations of functional groups and their occurrence in QM9S dataset and classification performance of our spectral classifier evaluated by Accuracy and F-1 Score.}
  \label{tab: fg-smarts}
  \centering
  \begin{tabular}{llllll}
    \toprule
    Functional groups     & SMARTS  & Count   & Accuracy & F-1  \\
    \midrule
    alkane       & [CX4;H3,H2,H1]                    & 123361 & 1.000 & 1.000 \\
    alkene       & [CX3]=[CX3]                       & 16382  & 0.994 & 0.977 \\
    alkyne       & [CX2]\#[CX2]                      & 16976  & 0.998 & 0.992 \\
    amine        & [NX3;!\$(NC=O)]                   & 38933  & 0.998 & 0.997 \\
    imine        & [NX2]=[CX3]                       & 15683  & 0.998 & 0.993 \\
    nitrile      & [NX1]\#[CX2]                      & 16356  & 0.998 & 0.990 \\
    alcohol      & [OX2H;!\$(OC=O)]                  & 42589  & 0.999 & 0.999 \\
    ether        & [OX2H0;!\$(OC=O);!\$([O]-[O])]    & 51295  & 0.995 & 0.994 \\
    haloalkane   & [\#6;!\$(C(=O)[F])][F]            & 2054   & 1.000 & 1.000 \\
    aldehyde     & [\#6,H][CX3H1](=O)                & 14828  & 0.997 & 0.986 \\
    ketone       & [\#6][CX3](=O)[\#6]               & 14690  & 0.995 & 0.977 \\
    ester        & [CX3](=O)[OX2H0]                  & 7011   & 0.994 & 0.947 \\
    amide        & [CX3](=O)[NX3]                    & 13541  & 0.998 & 0.992 \\
    arene        & [\$([cX2](:*):*),\$([cX3](:*):*)] & 21687  & 0.998 & 0.995 \\
    imidazole    & [\#7]:[\#6]:[\#7]                 & 6486   & 0.995 & 0.933 \\
    pyrazole     & [\#7]:[\#7]                       & 6927   & 0.998 & 0.982 \\
    oxazole      & [\#7]:[\#6]:[\#8]                 & 3706   & 0.998 & 0.961 \\
    isoxazole    & [\#7]:[\#8]                       & 3811   & 0.998 & 0.965 \\
    cyclopropane & C1CC1                             & 30386  & 0.986 & 0.971 \\
    epoxide      & C1OC1                             & 10200  & 0.993 & 0.954 \\

    \bottomrule
  \end{tabular}
\end{table}

\begin{table}[h]
  \caption{26 SMARTS representations of functional groups and their occurrence in QMe14S dataset and classification performance of our spectral classifier evaluated by Accuracy and F-1 Score.}
  \label{tab: fg-smarts-qme14s}
  \centering
  \begin{tabular}{lllll}
    \toprule
    Functional groups     & SMARTS &Count    & Accuracy & F-1  \\
    \midrule
    alkane         &[CX4;H3,H2,H1]                                 &158130  & 0.988 & 0.992 \\
    alkene         &[CX3]=[CX3]                                    &25795   & 0.976 & 0.940 \\
    enamine        &[CX3]=[CX3]([NX3])                             &2786    & 0.989 & 0.738 \\
    alkyne         &[CX2]\#[CX2]                                    &18678   & 0.996 & 0.956 \\
    amine          &[NX3;!\$(NC=O)]                                 &50406   & 0.984 & 0.968 \\
    imine          &[NX2]=[CX3]                                    &19815   & 0.976 & 0.862 \\
    nitrile        &[NX1]\#[CX2]                                    &19013   & 0.997 & 0.974 \\
    alcohol        &[OX2H;!\$(OC=O)]                                &51163   & 0.991 & 0.977 \\
    ether          &[OX2H0;!\$(OC=O);!\$([O]-[O])]                   &59375   & 0.964 & 0.907 \\
    aldehyde       &[\#6,H][CX3H1](=O)                              &16281   & 0.993 & 0.923 \\
    ketone         &[\#6][CX3](=O)[\#6]                              &16075   & 0.986 & 0.833 \\
    carboxylic acid&[CX3](=O)[OX2H1]                               &2227    & 0.998 & 0.958 \\
    ester          &[CX3](=O)[OX2H0]                               &10257   & 0.989 & 0.917 \\
    amide          &[CX3](=O)[NX3]                                 &16980   & 0.989 & 0.937 \\
    carbamate      &[OX2][CX3](=O)[NX3]                            &1673    & 1.000 & 1.000 \\
    arene          &[\$([cX2](:*):*),\$([cX3](:*):*)]                &31898   & 0.981 & 0.957 \\
    imidazole      &[\#7]:[\#6]:[\#7]                                 &8051    & 0.991 & 0.659 \\
    pyrazole       &[\#7]:[\#7]                                      &8804    & 0.989 & 0.860 \\
    oxazole        &[\#7]:[\#6]:[\#8]                                 &4024    & 0.997 & 0.680 \\
    isoxazole      &[\#7]:[\#8]                                      &4255    & 0.993 & 0.785 \\
    cyclopropane   &C1CC1                                          &31446   & 0.994 & 0.930 \\
    epoxide        &C1OC1                                          &10590   & 0.995 & 0.879 \\
    haloalkane     &[!\#1;!\$(C(=[O,S,Se])[\#9,\#17,\#35])][\#9,\#17,\#35] &19035   & 0.995 & 0.992 \\
    sulfide        &[SX2]                                          &8891    & 0.974 & 0.928 \\
    sulfone        &[SX4](=[O,Se])(=[O,Se])                        &3047    & 1.000 & 1.000 \\
    sulfonate      &[OX2][SX4](=[O,Se])(=[O,Se])                   &1663    & 0.998 & 0.967 \\
    \bottomrule
  \end{tabular}
\end{table}
\newpage

\subsection{Reconstruction performance}
As described in Section~\ref{sec: procedures}, our autoencoder continues to be updated during the training of the diffusion models. 
To evaluate its reconstruction quality, we assess the autoencoder on the held-out QM9S test set, comparing its performance both before and after joint training with the diffusion model.
Table~\ref{tab: recon} reports the mean squared error (MSE) between the input coordinates and the reconstructed geometries. 
We observe that the autoencoder achieves low reconstruction error after pretraining, indicating its capacity to encode and recover molecular geometries accurately. 
Moreover, the error further decreases after joint training, suggesting that the end-to-end optimization with the diffusion model enhances the quality of the learned latent representations.

\begin{table}[ht]
    \centering
    \caption{Autoencoder Reconstruction Performance}
    \begin{tabular}{ccc}
    \toprule
       & After AE training (\AA{}\textsuperscript{2})  & After DM training (\AA{}\textsuperscript{2}) \\
    \midrule
    MSE (test set) 
    & 3 $\times$ 10\textsuperscript{-4} & 6 $\times$ 10\textsuperscript{-6} \\
    \bottomrule
    \end{tabular}
    \label{tab: recon}
\end{table}

\subsection{Evaluation across multiple sampling runs}
\label{sec: multi-run}
We conduct repeated sampling experiments and report the variability in performance across three independent runs. Table~\ref{tab: repeat} summarises the mean and standard deviation of key evaluation metrics, including chemical graph similarity, spectral similarity (SIS and SIS\textsuperscript{*}), and molecular accuracy. All results are computed on stable sampled molecules, as described in Section~\ref{sec: exp-setting}.

\begin{table}[h]
  \caption{Evaluation of IR-GeoDiff trained on QM9S dataset over three repeated sampling runs. We report the mean and standard deviation for chemical graph similarity ($\mathrm{sim}_g$), Spectral Information Similarity (SIS), SIS\textsuperscript{*}, and molecular accuracy.}
  \label{tab: repeat}
  \centering
  \small
  \begin{tabular}{lcccccc}
    \toprule
    Method & $\mathrm{sim}_g$  & $\max \mathrm{sim}_g$   
    & $\mathrm{Mol\ acc}$(\%) & SIS 
    & $\max$ SIS & SIS\textsuperscript{*}\\
    \midrule

    ours  & \makecell{0.666$\pm$0.001} & \makecell{0.987$\pm$0.002} & \makecell{95.3$\pm$0.6} & \makecell{0.675$\pm$0.001}  & \makecell{0.947$\pm$0.004} & \makecell{0.718$\pm$0.001}\\
    
    \bottomrule
  \end{tabular}
\end{table}

\subsection{Evaluation under different sampling counts}
\label{sec: diff-sample}

\begin{figure}[h]
    \centering
    \includegraphics[width=0.6\linewidth]{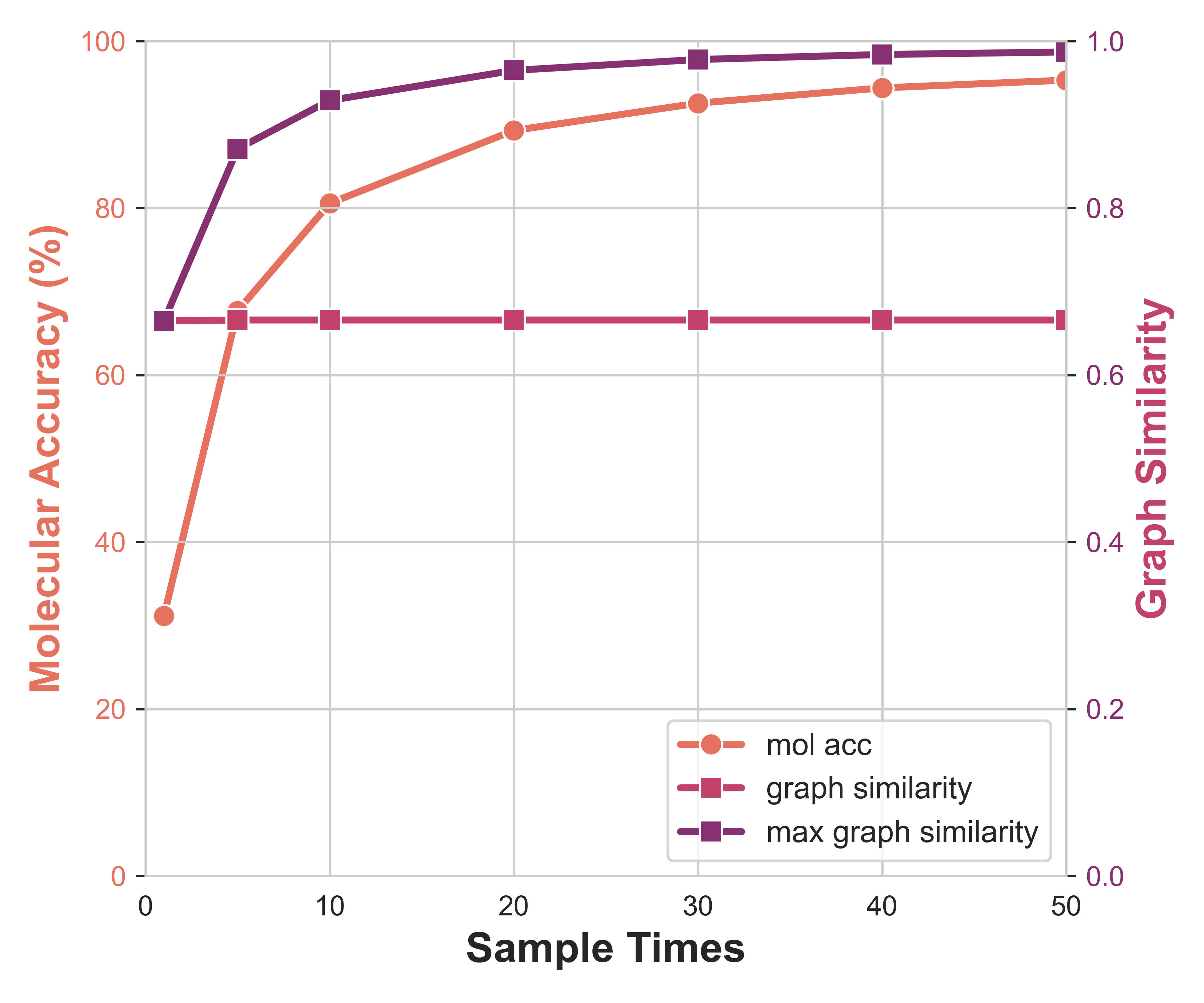}
    \caption{Effect of the number of samples per input spectrum on model performance, reporting molecular accuracy (mol acc), average graph similarity ($\mathrm{sim}_g$), and maximum graph similarity ($\max \mathrm{sim}_g$) across different sampling times.}
    \label{fig: diff-sample-times}
\end{figure}

We analyse the effect of varying the number of samples per input spectrum (1, 5, 10, 20, 30, 40, 50) on the QM9S dataset. As shown in Figure~\ref{fig: diff-sample-times}, the graph similarity $\mathrm{sim}_g$ remains stable across different sample sizes, whereas both $\max \mathrm{sim}_g$ and molecular accuracy consistently improve as the number of samples increases. The performance gains begin to saturate beyond 30 samples, indicating that using 50 samples provides a stable and representative evaluation of the model’s ability to recover the correct structure.

\subsection{Validity, stability and connectivity}
\label{sec: common_metrics}
We evaluate the quality of generated molecules using three common structural metrics: validity, stability, and connectivity.
We define validity as the proportion of generated molecules in which no atom exceeds its maximum allowed valence, and the overall molecular graph is connected, given that the atom types and counts are predefined.
Stability measures the proportion of molecules where all atoms are fully saturated up to their permitted valence limits.
Connectivity measures the proportion of molecules whose corresponding graphs are connected.
We report these metrics in Table~\ref{tab: common_metrics} for our full model and baseline models trained on the QM9S dataset and QMe14S dataset for reference, to provide a general assessment of molecular quality in the context of generative modeling. \textit{However, please note, these are not the primary evaluation metrics for our geometry recovery task, and are inappropriate to evaluate our approach.} Our main focus is on spectral and structural fidelity, as measured by SIS and $\mathrm{sim}_g$ in Table~\ref{tab: comparison} in the main paper, respectively.

\begin{table}[h]
  \caption{Comparison with baseline models on validity, stability, and connectivity.}
  \label{tab: common_metrics}
  \centering
  \begin{tabular}{llcccccc}
    \toprule
    Method & Dataset & \textcolor{gray}{Validity(\%)}  & \textcolor{gray}{Stability(\%)}   & \textcolor{gray}{Connectivity(\%)} \\
    \midrule
    {EDM~\citep{hoogeboom2022equivariant} } & \multirow{6}{6em}{QM9S}
    & \textcolor{gray}{99.04} & \textcolor{gray}{71.05} & \textcolor{gray}{97.81} \\
    {GEOLDM~\citep{xu2023geometric}} && \textcolor{gray}{99.08} & \textcolor{gray}{78.81} & \textcolor{gray}{98.67} \\
    GFMDiff~\citep{xu2024geometric} && \textcolor{gray}{99.38} & \textcolor{gray}{81.54} & \textcolor{gray}{98.09}\\
    EDM-variant && \textcolor{gray}{97.92} & \textcolor{gray}{89.48} & \textcolor{gray}{97.46} \\
    GEOLDM-variant && \textcolor{gray}{97.17} & \textcolor{gray}{84.19} & \textcolor{gray}{96.10}\\
    {ours} && \textcolor{gray}{98.53$\pm$0.10} & \textcolor{gray}{92.47$\pm$0.20} & \textcolor{gray}{98.35$\pm$0.10} \\
    \midrule
    EDM-variant & \multirow{3}{6em}{QMe14S}& \textcolor{gray}{97.45} & \textcolor{gray}{84.54} & \textcolor{gray}{95.31} \\
    GEOLDM-variant && \textcolor{gray}{98.14} & \textcolor{gray}{88.80} & \textcolor{gray}{96.98}\\
    {ours} && \textcolor{gray}{98.68$\pm$0.05} & \textcolor{gray}{92.56$\pm$0.11} & \textcolor{gray}{97.94$\pm$0.03} \\
    
    \bottomrule
  \end{tabular}
\end{table}


\subsection{Spectral Information Similarity}
\label{sec: sis}

The spectral information similarity (SIS) proposed by~McGill \etal\cite{mcgill2021predicting} are calculated to compare the spectra of sampled molecules and the input spectra.
First, each spectrum was normalized by summing all absorbance values to unity. We
then calculated the Spectral Information Divergence (SID), which measures the divergence
and peak overlap between any two spectra. Finally, SIS was computed based on SID as follows:
\begin{eqnarray}
        \text{SID}(Y_{\text{sampled}},Y_{\text{input}} )
        &=&
        \sum_i y_{\text{sampled,i}}
        \ln{\dfrac{y_{\text{sampled, i}}}{y_{\text{input,i}}}}
        + y_{\text{input,i}}\ln\dfrac{y_{\text{input,i}}}{y_{\text{sampled,i}}},
        \\
        \text{SIS}(Y_{\text{sampled}},Y_{\text{input}} )
        &=& \dfrac{1}{1+\text{SID}(Y_{\text{sampled}},Y_{\text{input}} )},
\end{eqnarray}
where $y_{\text{sampled}}$ and $y_{\text{input}}$ are vectors representing the spectra of sampled molecules~$Y_{\text{sampled}}$ and the input spectra~$Y_{\text{input}}$, respectively.

\subsection{Spectral–atom cross-attention map}
\label{sec: ca-hydrogen}
\begin{figure}[ht]
    \centering
    \includegraphics[width=1.0\linewidth]{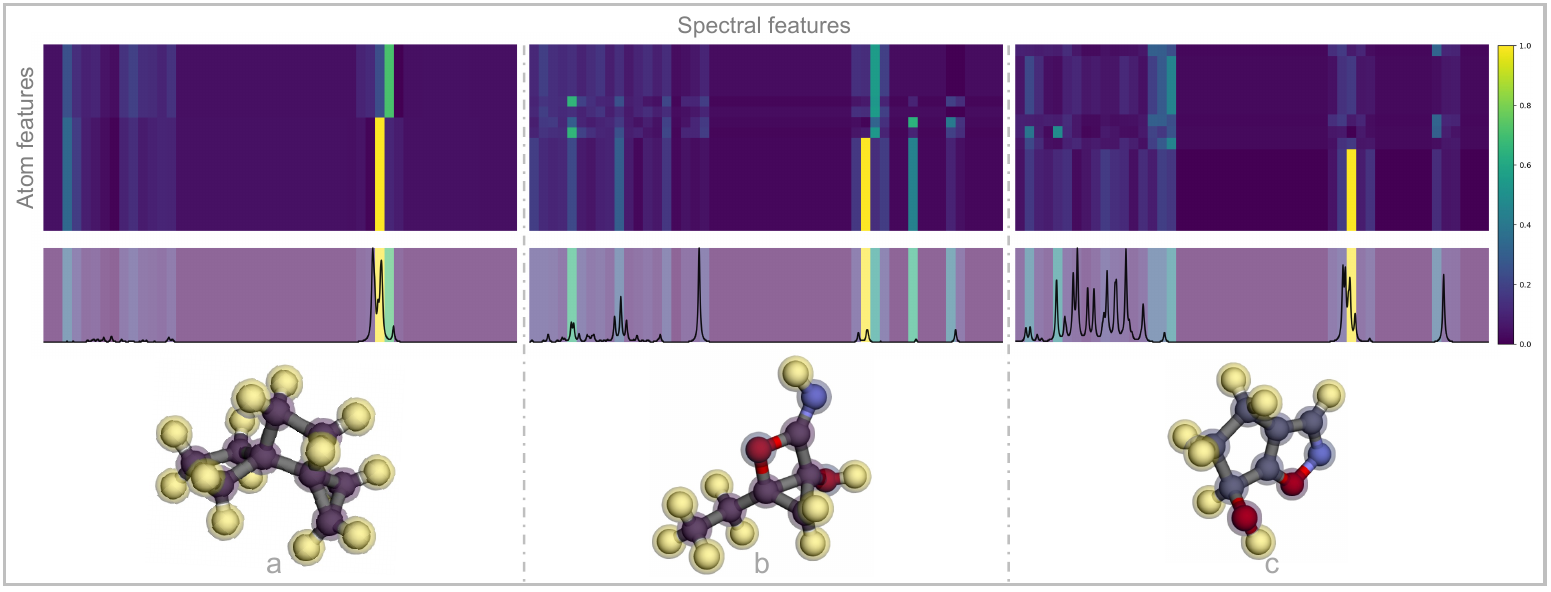}
    \caption{Spectral–atom cross-attention emphasizing the hydrogen atoms.}
    \label{fig: hydro-attn}
\end{figure}

As discussed in Section~\ref{sec: attn}, the model adaptively allocates attention across different atom types based on the spectral features. 
While Figure~\ref{fig: attn_map}B and Figure~\ref{fig: attn_map}C highlight the model's focus on non-carbon heavy atoms and the global carbon framework, 
Figure~\ref{fig: hydro-attn} further demonstrates that the model is also capable of attending strongly to hydrogen atoms throughout the molecule. 
This observation reinforces that the model does not follow a fixed atom-specific bias but instead dynamically adjusts its attention based on the spectral context.

\section{Results of Vibrational Analysis by Gaussian 16}
\label{sec: gaussian-output}
\begin{figure}[h]
    \centering
   \includegraphics[width=0.67\linewidth]{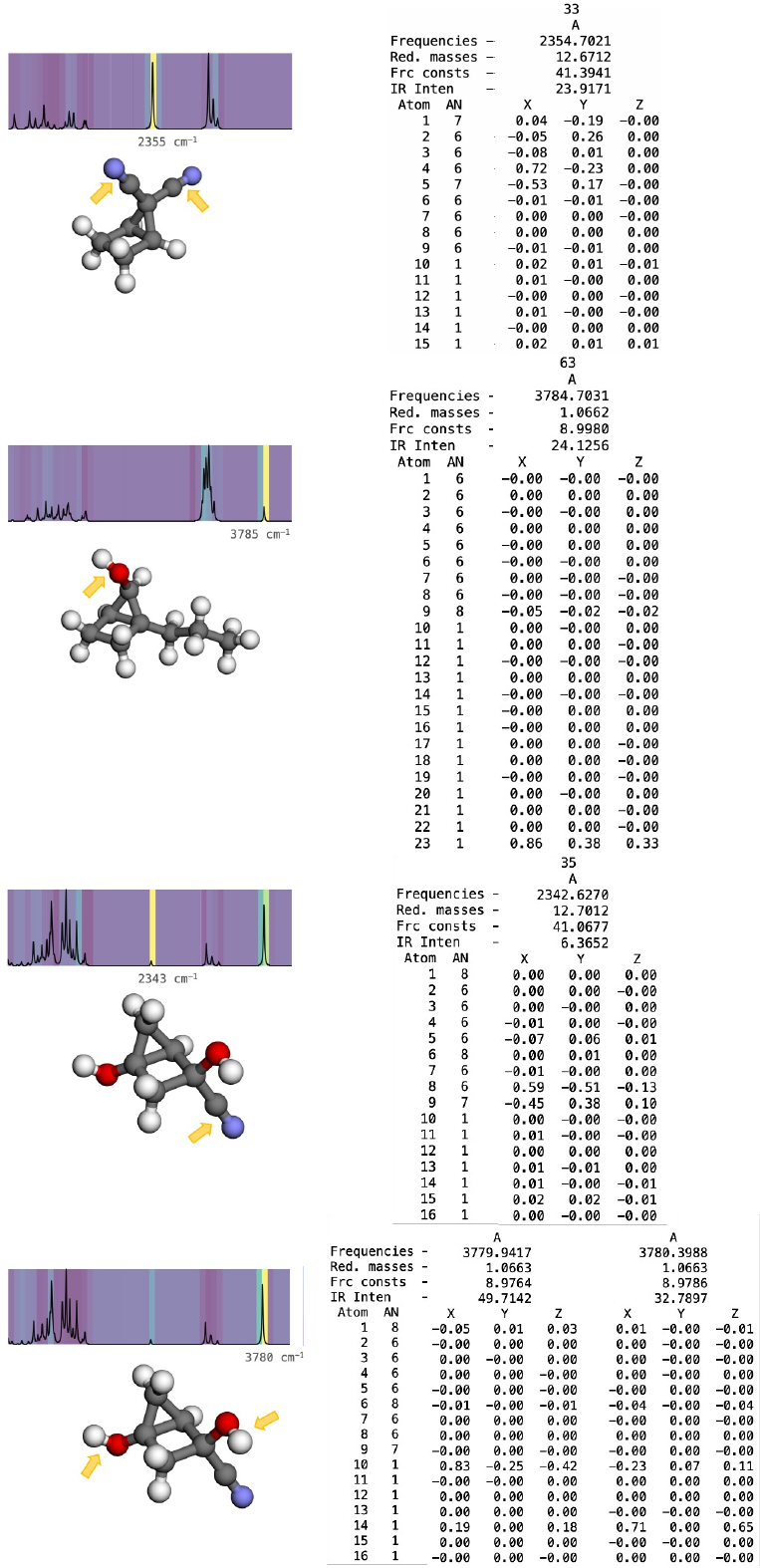}
    \caption{Attention map and quantum mechanical vibrational analysis. \textbf{Left}: Cross-attention scores from the model between spectral features and edge features. (For atoms, hydrogen: {white}~\hydrogendot, \textcolor{gray}{carbon: {gray}}~\atomdot{darkgray}, \textcolor{red}{oxygen: {red}~\atomdot{red}}, \textcolor{Periwinkle}{nitrogen: {blue}~\atomdot{Periwinkle}}.) \textbf{Right}: Gaussian 16 vibrational analysis output. The displacement vectors (x, y, z) indicate atomic movements in each vibrational mode.}
    \label{fig: gaussian_output}
\end{figure}

Figure~\ref{fig: gaussian_output} shows excerpts from the vibrational analysis output of Gaussian 16 package~\citep{g16}. The vibrational modes shown on the right correspond to the spectral peaks highlighted on the left. Each table presents displacement vectors (x, y, z) for all atoms in the molecule, indicating how atomic positions change during the corresponding vibrational mode. These modes correspond to spectral peaks that receive high attention from our model, further validating its interpretability.

\newpage
\bibliography{iclr2026_conference}
\bibliographystyle{iclr2026_conference}

\end{document}

%% file: math_commands.tex
\usepackage{amsmath,amsfonts,bm}

\def\eqref#1{equation~\ref{#1}}

\def\1{\bm{1}}

\DeclareMathAlphabet{\mathsfit}{\encodingdefault}{\sfdefault}{m}{sl}
\SetMathAlphabet{\mathsfit}{bold}{\encodingdefault}{\sfdefault}{bx}{n}

